\documentclass[sigconf]{acmart}

\usepackage{enumitem}
\usepackage{multirow}
\usepackage{pifont}

\usepackage[version=4]{mhchem}

\usepackage{float}
\usepackage{lipsum}
\usepackage{multicol}
\usepackage{siunitx}
\usepackage{tabularx,booktabs}
\usepackage{etoolbox}
\usepackage{newtxtext}
\usepackage{newtxmath}
\usepackage{booktabs}
\usepackage{threeparttable}

\usepackage{caption, subcaption, overpic, textpos}

\definecolor{LightCyan}{rgb}{0.88,1,1}
\definecolor{LightRed}{rgb}{1,0.88,1}
\definecolor{LightYellow}{rgb}{1,1,0.88}
\definecolor{Grey}{rgb}{0.75,0.75,0.75}
\definecolor{DarkGrey}{rgb}{0.55,0.55,0.55}
\definecolor{DarkGreen}{rgb}{0,0.65,0}

\newlength\savewidth

\definecolor{baselinecolor}{gray}{.9}

\newcolumntype{x}[1]{>{\centering\arraybackslash}p{#1pt}}
\newcolumntype{y}[1]{>{\raggedright\arraybackslash}p{#1pt}}
\newcolumntype{z}[1]{>{\raggedleft\arraybackslash}p{#1pt}}

\newcommand{\tianhong}[1]{\textcolor{blue}{\ignorespaces}}
\newcommand{\huiwen}[1]{\textcolor{red}{\ignorespaces}}
\newcommand{\han}[1]{\textcolor{cyan}{\ignorespaces}}
\newcommand{\dilip}[1]{\textcolor{orange}{\ignorespaces}}
\newcommand{\shlok}[1]{\textcolor{green}{\ignorespaces}}

\colorlet{darkgreen}{green!65!black}
\colorlet{darkblue}{blue!75!black}
\colorlet{darkred}{red!80!black}
\definecolor{lightblue}{HTML}{0071bc}
\definecolor{lightgreen}{HTML}{39b54a}

\AtBeginDocument{%
  }

\copyrightyear{2026}
\acmYear{2026}
\setcopyright{cc}
\setcctype{by}
\acmConference[WWW '26] {Proceedings of the ACM Web Conference 2026}{April 13--17, 2026}{Dubai, United Arab Emirates.}
\acmBooktitle{Proceedings of the ACM Web Conference 2026 (WWW '26), April 13--17, 2026, Dubai, United Arab Emirates}
\acmISBN{979-8-4007-2307-0/2026/04}
\acmDOI{10.1145/3774904.3792515}
\begin{document}

\title{Hi-GMAE: Hierarchical Graph Masked Autoencoders}

\author{Chuang Liu}
% \authornote{Both authors contributed equally to this work.}
% \author{Zelin Yao}
% \email{zelinyao@whu.edu.cn}
\affiliation{%
  \institution{Sangfor Technologies Inc.}
  \city{Shenzhen}
  \country{China}
}
\email{chuangliu@whu.edu.cn}

\author{Zelin Yao}
\affiliation{%
  \institution{Wuhan University}
  \city{Wuhan}
  \country{China}
  }
\email{zelinyao@whu.edu.cn}

\author{Xueqi Ma}
\affiliation{%
  \institution{The University of Melbourne}
  \city{Melbourne}
  \country{Australia}
}
\email{xueqim@student.unimelb.edu.au}

\author{Mukun Chen}
\affiliation{%
  \institution{Wuhan University}
  \city{Wuhan}
  \country{China}
  }
\email{cmk0910@whu.edu.cn}

\author{Luzhi Wang}
\affiliation{%
 \institution{Dalian Maritime University}
 \city{Dalian}
 \country{China}
 }
 \email{wangluzhi0@gmail.com}

\author{Jia Wu}
\affiliation{%
  \institution{Macquarie University}
  \city{Sydney}
  \country{Australia}
  }
\email{jia.wu@mq.edu.au}

\author{Wenbin Hu}
\authornote{Corresponding Author}
\affiliation{%
  \institution{Wuhan University, Wuhan University Shenzhen Research Institute}
  \city{Wuhan}
  \country{China}
}
\email{hwb@whu.edu.cn}

\renewcommand{\shortauthors}{Chuang Liu et al.}

\begin{abstract}
Graph Masked Autoencoders (GMAEs) have emerged as a notable self-supervised learning approach for graph-structured data. Existing GMAE models primarily focus on reconstructing node-level information, categorizing them as single-scale GMAEs. This methodology, while effective in certain contexts, tends to overlook the complex hierarchical structures inherent in many real-world graphs.  For instance, molecular graphs exhibit a clear hierarchical organization in the form of the \textit{atoms-functional groups-molecules} structure. Therefore, the inability of single-scale GMAE models to incorporate these hierarchical relationships often results in an inadequate capture of crucial high-level graph information, leading to a noticeable decline in performance. To address this limitation, we propose Hierarchical Graph Masked AutoEncoders (Hi-GMAE), a novel multi-scale GMAE framework designed to handle the hierarchical structures within graphs. First, Hi-GMAE constructs a multi-scale graph hierarchy through graph pooling, enabling the exploration of graph structures across different granularity levels. To ensure masking uniformity of subgraphs across these scales, we propose a novel coarse-to-fine strategy that initiates masking at the coarsest scale and progressively back-projects the mask to finer scales. Furthermore, we integrate a gradual recovery strategy with the masking process to mitigate the learning challenges posed by completely masked subgraphs. Diverging from the standard graph neural network (GNN) used in GMAE models, Hi-GMAE modifies its encoder and decoder into hierarchical structures. This entails using GNN at the finer scales for detailed local graph analysis and employing a graph transformer at coarser scales to capture global information. Such a design enables Hi-GMAE to effectively capture the multi-level information inherent in complex graph structures. Our experiments on 17 graph datasets, covering two graph learning tasks, consistently demonstrate that Hi-GMAE outperforms 29 state-of-the-art self-supervised competitors in capturing comprehensive graph information.  Codes are available at \url{https://github.com/LiuChuang0059/Hi-GMAE}.
\end{abstract}





\maketitle

\section{Introduction}
\label{sec:introduction}

\begin{figure}[!t] % !ht% \setlength{\abovecaptionskip}{-0.1cm}   %调整图片标题与图距离
\begin{center}
\includegraphics[width=1.0\linewidth]{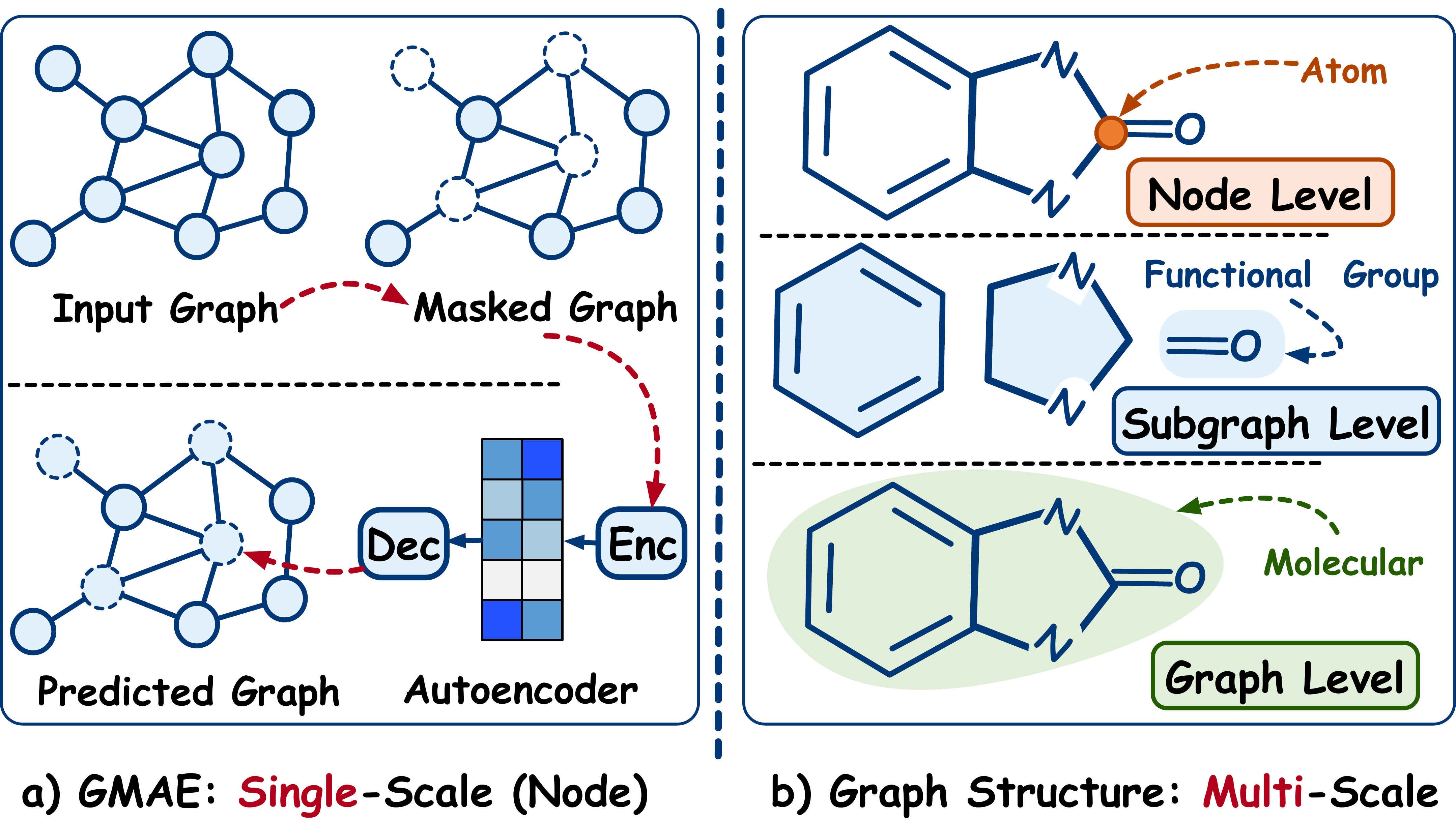}
\end{center}
\caption{
%GMAE Models \textbf{vs.} Graph Structure.  
\textbf{a)} The GMAE models focus on reconstructing randomly masked nodes using an autoencoder architecture, with a primary focus on the node level, categorizing them as  {\color[HTML]{A40203} \textbf{single-scale}} approaches. \textbf{b)} The inherent hierarchical structure of a molecule graph, which contains node, subgraph, and overall graph level information, is termed as {\color[HTML]{A40203} \textbf{multi-scale}}.}
\label{fig:motiv}
\end{figure}

In many real-world graphs, especially within the academic, social, and biological sectors, the availability of labeled data is often limited. This poses a significant challenge for traditional supervised learning approaches in graph analysis. To address this challenge, Graph Self-supervised Pre-training (GSP) methods have emerged as a powerful tool~\cite{survey-pretraining,cpt-hg,kpgt,gppt,mask-kg,w2pgnn,bandana}, utilizing the inherent graph data structures and properties to extract meaningful representations even in label-scarce environments.

Current GSP methods can be categorized into two primary streams: \textbf{1) Contrastive} GSP methods, such as GraphCL~\cite{graphCL} and SimGRACE~\cite{simGrace}, involve  constructing multiple graph views through augmentation and learning representations by contrasting positive and negative samples; \textbf{2) Generative} GSP methods, such as GraphMAE~\cite{graphmae} and MaskGAE~\cite{maskgae}, emphasize learning node representations through a reconstruction objective. Generative GSP methods have been shown to be simpler and more effective compared to contrastive approaches that require carefully designed augmentation and sampling strategies~\cite{graphmae2}. Accordingly, this paper explores the capabilities of generative GSP methods, specifically graph masked autoencoders (GMAEs), and further examines their potential in graph learning tasks.

% as evidenced in other fields.
% The efficacy of generative methods is further underscored by the enormous successes of models such as BERT and ChatGPT in Natural Language Processing (NLP)~\cite{bert} and MAE in Computer Vision (CV)~\cite{mae}. %上面这句话感觉没有信息量，下面一句也是
% These successes highlight the significant potential of generative approaches across various domains.

Current GMAE models, as illustrated in Figure~\ref{fig:motiv} (a), primarily focus on reconstructing node features, capturing mainly node-level information. However, this approach, categorized as  \textbf{single-scale}, is inconsistent with the  \textbf{multi-scale} structure inherent in graphs, such as social networks, recommendation networks, and molecular graphs, which is widely recognized as a cornerstone of  graph representation learning~\cite{sagpool,diffpool,pooling-survey}. For example, as displayed in the molecular graph in Figure~\ref{fig:motiv} (b), in addition to node-level atom information, a subgraph-level of information exists within functional groups (\textit{i.e.}, benzene rings). This multi-scale nature of the graph enables a richer representation of molecular characteristics, capturing complex interactions and relationships that exceed node-level information. However, current GMAE models face limitations in learning this complex, high-level structural information, resulting in a decline in performance. Therefore, this leads to a question: \textit{how can we effectively capture the hierarchical information inherent in graphs? }

% while GMAE models offer significant insights at the node level, they overlook the graph's innate multi-scale structure, indicating the potential for more profound and comprehensive learning.
% Recently, some graph learning methods have incorporated the mult-scale analysis techniques. For example, :

% Recognizing and leveraging this hierarchical complexity could lead to more accurate representations in graph-based molecular modeling. 

% potentially overlooking broader structural patterns and hierarchies within the graph and leading to sub-optimal representations.
% the primary focus of current GMAE models lies at the node level, categorizing it as a {\color[HTML]{A40203} \textbf{single-scale}} approach. In this setup, the model primarily concentrates on individual node attributes and their immediate connections,  failing to capture high-level structural information. 

% For example, if a molecule is represented by averaging its atoms, important infor- mation about functional groups like benzene rings or hydroxides would be completely disregarded.

% encounters limitations in learning the complex, high-level structural information inherent in graphs.  Contrastingly, recent research acknowledges the {\color[HTML]{A40203} \textbf{multi-scale}} or hierarchical structure as the gold standard for graph representation learning~\cite{sagpool,diffpool,pooling-survey}. 

To tackle this challenge, we propose the Hierarchical Graph Masked AutoEncoders (Hi-GMAE).  The Hi-GMAE framework comprises three principal components: \textbf{1) Multi-scale Coarsening:} The first step in extracting hierarchical information from graphs involves constructing a series of coarse graphs at multiple scales. These scales range from fine- to coarse-grained graphs. Specifically, we utilize graph pooling methods that progressively cluster nodes into super-nodes, thereby generating coarsened graphs. \textbf{2) Coarse-to-Fine (CoFi) Masking with Recovery:} Building on the multi-scale coarse graphs, we introduce a novel masking strategy. This strategy is designed to maintain the consistency of masked subgraphs across all scales. The process commences with applying random masking to the coarsest graph. The mask matrix is then back-projected to the finer scales using an unpooling operation. This CoFi approach ensures seamless and consistent masking across various levels of the graph hierarchy. Additionally, to address the challenges of learning from initially fully masked subgraphs, we introduce a gradual recovery strategy that selectively unmasks certain nodes. \textbf{3) Fine- and Coarse-Grained (Fi-Co) Encoder and Decoder:}  To effectively capture diverse information at different graph scales, we develop a hierarchical encoder that combines both fine-grained graph convolution and coarse-grained graph transformer (GT) modules. The fine-grained graph convolution modules capture local information in low-level graphs, while the coarse-grained GT modules focus on global information in high-level graphs. Correspondingly, a symmetrical and relatively lightweight decoder model is employed. This decoder is tasked with gradually reconstructing and projecting the representations, obtained from the encoder, back to the original graph scale. The hierarchical nature of the encoder and decoder is crucial for comprehensively capturing and representing the multi-level structural information observed in graphs. In summary, Hi-GMAE provides a self-supervised framework for effectively understanding and leveraging hierarchical graph information. By integrating multi-scale coarsening, an innovative masking strategy, and a hierarchical encoder-decoder architecture, Hi-GMAE effectively captures the graph structures at various levels.

% to facilitate the learning process, especially during initial epochs, we incorporate a node recovery strategy. This involves selectively unmasking certain nodes, with the number of recovered nodes gradually decreasing over the training process.

To evaluate the effectiveness of the Hi-GMAE model, we conduct comprehensive experiments on a range of widely-used datasets. The experiments include two graph learning tasks: unsupervised learning and transfer learning. The results demonstrate that Hi-GMAE outperforms existing state-of-the-art models in both contrastive and generative pre-training domains. The improvements highlight the advantages of our multi-scale GMAE approach over conventional single-scale ones. Our main contributions are summarized as follows:

\begin{enumerate}[leftmargin=12pt]
  \item  We introduce Hi-GMAE, a novel self-supervised graph pre-training framework, which incorporates several carefully designed key techniques, including multi-scale coarsening, an innovative masking strategy, and a hierarchical encoder-decoder architecture, to effectively capture multi-scale graph information.

  \item We evaluate Hi-GMAE's performance through extensive experiments, comparing its efficacy with generative and contrastive baselines across three graph tasks on various real-world graph datasets. The experimental results consistently validate the effectiveness of Hi-GMAE.
\end{enumerate}

% 1. We explore the masking strategies in MIM and demon- strate that the previous random sampling method is sub- optimal. We find that the pre-training results can be sig- nificantly improved by slightly raising the masking rates of the informative foreground image patches.

% 2. It is worth noting that the proposed  is a plug-and- play module for GMAE frameworks and could be readily deployed into GMAE methods, such as GraphMAE and MaskGAE.

% As a plug-and-play module, 

\section{Related Work}
\label{sec:related work}

\textbf{Graph Self-supervised Pre-training.} 
Inspired by the success of pre-trained language models, such as BERT~\cite{bert} and ChatGPT~\cite{gpt}, numerous efforts have been directed towards studying GSP. GSP is divided into contrastive and generative methods based on model architecture and objective design.  In the past few years, \textbf{contrastive} methods have dominated the field of graph representation learning~\cite{dgi,nodemixup,graphCL,costa}. Its success is largely due to elaborate data augmentation designs, negative sampling, and contrastive loss. In contrast, \textbf{generative} methods focus on recovering the missing parts of the input data. For example, GAE~\cite{gae} is a method that reconstructs the adjacency matrix. Multiple GAE variants utilize graph reconstruction to pre-train Graph Neural Networks (GNNs), including  SeeGera~\cite{seegera} and SIGVAE~\cite{siggae}. Recently, a paradigm shift towards GMAEs has displayed promising results in various tasks. GMAEs mainly focus on reconstructing the randomly masked input contents (\textit{e.g.}, nodes and edges) using autoencoder architecture. A prominent and innovative example, GraphMAE~\cite{graphmae},  reconstructs masked node features, introducing novel techniques such as re-mask decoding and scaled cosine error.  In addition, approaches like MaskGAE~\cite{maskgae}, S2GAE~\cite{s2gae}, and GiGaMAE~\cite{gigamae} jointly reconstruct masked edges and nodes. Furthermore, the integration of contrastive learning with GMAEs, such as SimSGT~\cite{simsgt}, lrGAE~\cite{lrGAE}, and GCMAE~\cite{GCMAE,gcmae2}, and application of self-distillation in RARE~\cite{rare} further diversify the methodological landscape, enhancing GMAEs' effectiveness. Unlike the above mentioned GMAE methods, which primarily adopt a random masking strategy, StructMAE~\cite{structmae} proposes a novel structure-guided masking strategy. GMAEs are applied across various domains, including heterogeneous graph representation~\cite{hgmae,RMR}, protein surface prediction~\cite{protein-mae}, and action recognition~\cite{skeletonmae}. However, all the aforementioned GMAE methods overlook the hierarchical information inherent in graphs, potentially inhibiting the model's capabilities.

\noindent  \textbf{Hierarchical Graph Learning.}
In recent years, hierarchical graph learning has emerged as a pivotal study area within the realm of GNNs. This approach is characterized by its capacity to capture multi-scale structural information, thereby enhancing the effectiveness of learning from graph data. Innovations such as DiffPool~\cite{diffpool}, Graph U-Nets~\cite{graph-u-net}, SAGPool~\cite{sagpool}, and EigenPool~\cite{eigenpool} have been crucial in laying the groundwork for hierarchical representation learning in graphs. These approaches involve strategically reducing graphs into smaller, more condensed subgraphs through pooling, ensuring the preservation of essential structural details. Moreover, further advancements have been achieved with the introduction of hierarchical GT, such as ANS-GT~\cite{ans-gt} and HSGT~\cite{hsgt}, thereby scaling and enhancing GT learning. Exploring hierarchical information has also extended to specialized domains, notably the enhancement of molecular~\cite{himol} and protein~\cite{high-ppi} representations. These advancements emphasize the potential of hierarchical graph models in handling complex data. Despite this progress, the majority of existing research focuses on supervised learning, with limited attention paid to integrating hierarchical characteristics with GSP, especially concerning generative GSP models. Our work seeks to fully exploit the potential of hierarchical structures in a self-supervised manner.

\section{Preliminaries}
\label{sec:preliminary}

\textbf{Notations.} A graph $\mathcal{G} =(\mathcal{V}, \mathcal{E}) $ can be represented by an adjacency matrix $\boldsymbol{A} \in \{0, 1\}^{ n \times n}$ and a node feature matrix $\boldsymbol{X} \in \mathbb{R}^{ n \times d_0}$, where $\mathcal{V}$ denotes the node sets, $\mathcal{E}$ denotes the edge sets,  $n$ is the number of nodes, and $d_0$ is node feature dimensions. Furthermore, $\boldsymbol{A}[i, j]=1$ if an edge between nodes $v_{i}$ and $v_{j}$ exists; otherwise, $\boldsymbol{A}[i, j]=0$.

\noindent \textbf{Graph Masked Autoencoders.} GMAEs operate as a self-supervised pre-training framework, which focuses on recovering masked node features or edges based on the unmasked node representations. We explore node feature reconstruction as an illustrative example to clarify the GMAE methods. GMAEs comprise two essential components: an encoder ($f_{E}(\cdot)$) and a decoder ($f_{D}(\cdot)$). The encoder maps each unmasked node $v_i \in \mathcal{V}_{u}$ to a $d$-dimensional vector $\boldsymbol{h}_i \in \mathbb{R}^{d}$, with $\mathcal{V}_{u}$ representing the unmasked node set, while the decoder reconstructs the masked node features from these vectors. The entire process can be formally represented as:
\begin{equation}
\boldsymbol{H}_{u}=f_E(\boldsymbol{A}, \boldsymbol{X}_{u}); \quad \hat{\boldsymbol{X}}=f_D(\boldsymbol{A}, \boldsymbol{H}_{u}),
\end{equation}
where $\boldsymbol{X}_{u}$ and $\boldsymbol{H}_{u}$ denote unmasked nodes' input features and encoded embeddings, respectively, and  $\hat{\boldsymbol{X}}$ represents the reconstructed features of all nodes. In addition, GMAEs optimize the model by minimizing the discrepancy between the reconstructed representation of the masked nodes, $\hat{\boldsymbol{X}}_{m} \subset \hat{\boldsymbol{X}}$,  and their original features, $\boldsymbol{X}_{m} \subset \boldsymbol{X}$.

\noindent \textbf{Graph Pooling and Unpooling.}
Graph pooling generates coarse graphs that retain essential information while reducing the number of nodes. This process involves clustering nodes into super-nodes. Given a graph $\mathcal{G} =(\mathcal{V}, \mathcal{E}) $, the resulting coarse graph is smaller $\mathcal{G}^{\prime}=\left(\mathcal{V}^{\prime}, \mathcal{E}^{\prime}\right)$. This is achieved by partitioning  $\mathcal{V}$  into disjoint clusters $\left\{\mathcal{C}_1, \mathcal{C}_2, \cdots, \mathcal{C}_{n^{\prime}}\right\}$, where each cluster $\mathcal{C}_i$ represents a super-node in the coarse graph $\mathcal{G}^{\prime}$, and $n^{\prime} = \| \mathcal{V}^{\prime} \|$. The aforementioned pooling process can be further delineated through a cluster assignment matrix $\boldsymbol{P} \in\{0,1\}^{n \times n^{\prime}}$, where $\boldsymbol{P}_{i j}=1$ if node $v_i$ belongs to cluster $\mathcal{C}_j$. Then, the new feature and adjacency matrices of $\mathcal{G}^{\prime}$ are defined as follows: $\boldsymbol{X}^{\prime} = \boldsymbol{P}^{\top} \boldsymbol{X} \in \mathbb{R}^{ n^{\prime} \times d_0}$, and $\boldsymbol{A}^{\prime} = \boldsymbol{P}^{\top} \boldsymbol{A} \boldsymbol{P} \in \mathbb{R}^{ n^{\prime} \times n^{\prime}}$. Following the pooling process, the number of nodes in $\mathcal{G}^{\prime}$ are significantly reduced compared to $\mathcal{G}$. This reduction is quantified by the pooling ratio, defined as $r_{p}=n^{\prime}/n$.  In contrast, the unpooling operations within the encoder-decoder architecture up-sample the coarse representations back to their original resolution. Formally, the unpooling operation can be expressed as: $\boldsymbol{X} = \boldsymbol{P} \boldsymbol{X}^{\prime} \in \mathbb{R}^{ n \times d_0}$.

\section{Proposed Method}
\label{sec:propose-method}

This section describes Hi-GMAE's model architecture. First, we elaborate on the construction of hierarchical graphs (\textbf{$\S$\ref{sec:generate-coarse}}). Then, we introduce the coarse-to-fine masking module (\textbf{$\S$\ref{sec:masking}}). Finally, a comprehensive exploration of the fine- and coarse-grained encoder and decoder is provided (\textbf{$\S$\ref{sec:encoder}}).  

%Finally, the complexity analysis of Hi-GMAE is clarified  (\textbf{$\S$\ref{sec:complexity}}).

% Additionally, the training and inference details of Hi-GMAE are outlined (\textbf{$\S$\ref{sec:overall-arch}}). 

\begin{figure}[!t] % !htb
\begin{center}
\includegraphics[width=1.0\linewidth]{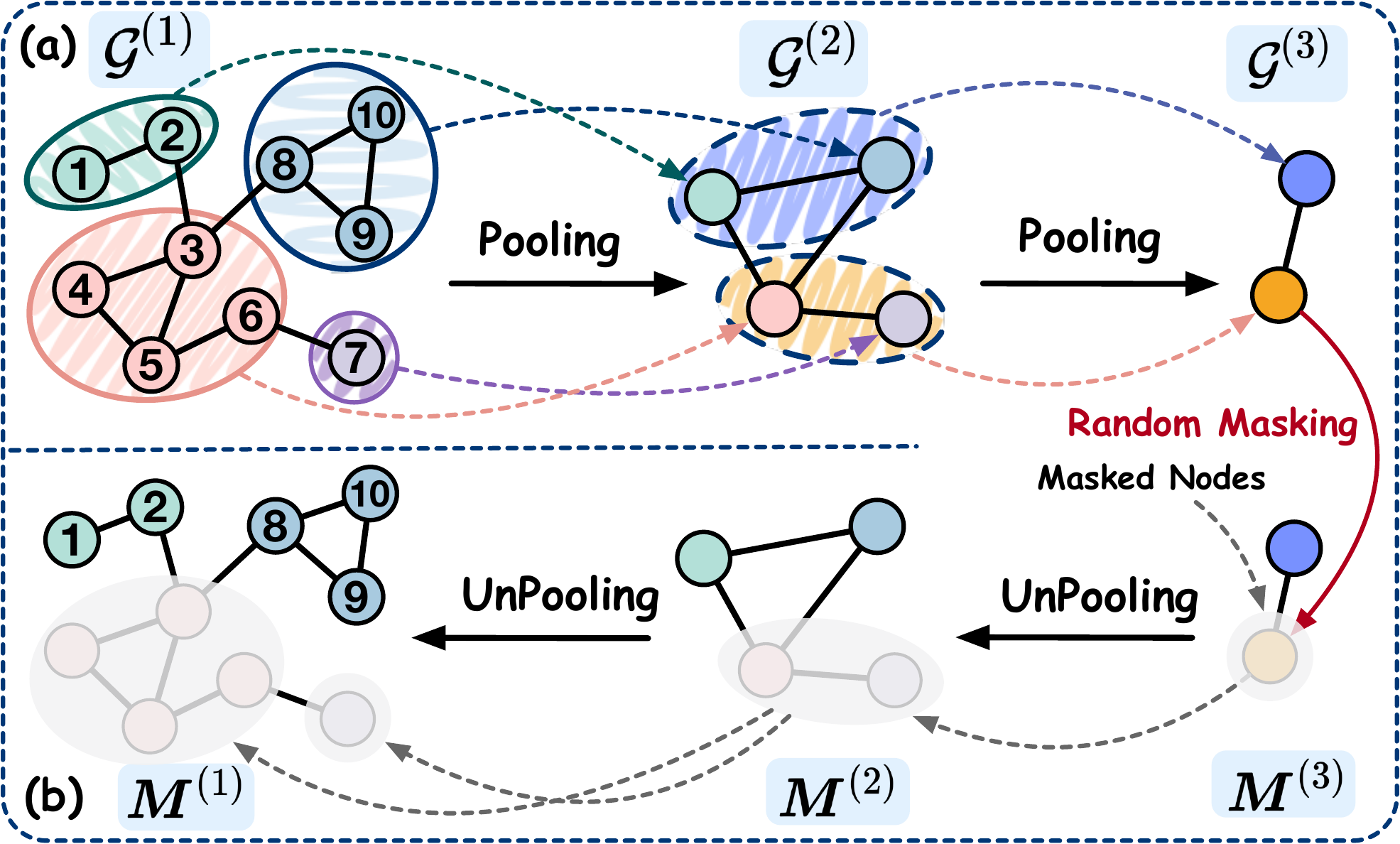}
\end{center}
\caption{Illustration of multi-scale graph generation and masking matrix construction.  \textbf{(a) Generating Coarse Graphs:} We utilize the graph pooling algorithm, which clusters nodes into super-nodes, to generate the coarse graph. \textbf{(b) CoFi Masking:} We conduct random masking on the coarsest graph and then project the mask matrix onto the fine-grained graph through an unpooling process. This approach ensures that the masking pattern remains consistently aligned across different scales 
 }
\label{fig:masking}
\end{figure}

%of the graph. Please note that the above two processes, including pooling and unpooling, are iterated across $S$ scales, although only three scales are presented here for the sake of clarity.

\subsection{Generating Coarse Graphs}
\label{sec:generate-coarse}
To capture high-level context information within graph structures, our approach involves constructing a series of progressively coarse graphs $\{ \mathcal{G}^{(l)} \}_{l=1}^S$ from the original graph $\mathcal{G}^{(1)} = \mathcal{G}$, as illustrated in Figure~\ref{fig:masking} (a). Each coarse graph $\mathcal{G}^{(l)}$ represents the graph's structure at the $l$-th coarse level. The depth of this hierarchical structure $S$ and the pooling ratio $r_p$ are predefined hyperparameters. Our framework allows flexibility in selecting the graph pooling algorithm, and we choose METIS~\cite{metis} for its efficiency and scalability. This algorithm requires only a single computation during the pre-processing phase. The pooling process from the $(l-1)$-th scale to the $l$-th is mathematically represented as follows:
\begin{equation}
    \boldsymbol{A}^{(l)} = \left(\boldsymbol{P}^{(l-1)}\right)^{\top} \boldsymbol{A}^{(l-1)} \boldsymbol{P}^{(l-1)} \in \mathbb{R}^{ n^{(l)} \times n^{(l)}},
\end{equation}
where $\boldsymbol{A}^{(l)}$ represents the coarse graph's adjacency matrix, and $\boldsymbol{A}^{(1)} = \boldsymbol{A}$. Through iterative pooling, we obtain the  $S$ scale adjacency matrices $\{ \boldsymbol{A}^{(l)} \}_{l=1}^S$ for the coarse graphs at every hierarchical level.

\subsection{Coarse-to-Fine Masking with Recovery}
\label{sec:masking}

\subsubsection{Coarse-to-Fine Masking}

Traditional random masking strategies used in previous GMAE models, though effective in single-scale graphs, fail to ensure consistent masked information across multiple scales in hierarchical graph representation learning. To address this challenge, we propose a CoFi masking strategy to manage graph data within the hierarchical encoder, ensuring the consistency of unmasked subgraphs across different scales. Utilizing the coarse graph set $\{ \mathcal{G}^{(l)} \}_{l=1}^S$ generated in \textbf{$\S$\ref{sec:generate-coarse}}, we first apply random masking to the coarsest graph $\mathcal{G}^{(S)}$ with a masking matrix $\boldsymbol{M}^{(S)} \in \{0,1\}^{ n^{(S)} \times 1}$, where $n^{(S)}$ denotes the number of nodes in the $S$-th scale. This process generates a new version of the graph where a subset of nodes is masked (\textit{i.e.}, the masked nodes' features are set to zero). Then, the mask matrix $\boldsymbol{M}^{(S)}$ is iteratively back-projected to the lower scales, as depicted in Figure~\ref{fig:masking} (b). This back-projection is achieved through the unpooling operation, formulated as:
\begin{equation}
    \boldsymbol{M}^{(S-1)} = \left(\boldsymbol{P}^{(S-1)}\right)^{\top} \boldsymbol{M}^{(S)} \in \{0,1\}^{ n^{(S-1)} \times 1} .
\end{equation}
By applying this recursive back-projection method, we obtain the mask matrix for all scales in the series, denoted as $\{ \boldsymbol{M}^{(l)}\}_{l=1}^S$. This design ensures the consistency of unmasked, visible subgraphs across all the scales.

% Specifically, a subset of nodes $\mathcal{V}_{\text{mask}} \subset \mathcal{V}$ is sampled for masking. According to the methodology described in~\cite{graphmae}, the features of these sampled nodes are replaced with a learnable vector $\boldsymbol{x}_{[\text{M}]} \in \mathbb{R}^d$. Accordingly, for a node $v_i$ within the masked node subset $\mathcal{V}_{\text{mask}}$, its feature $\widetilde{\boldsymbol{x}}_i$ in the altered feature matrix $\widetilde{\boldsymbol{X}}$ is defined as follows:
% \begin{equation}
% \widetilde{\boldsymbol{x}}_i= \begin{cases}\boldsymbol{x}_{[\text{M}]} & v_i \in \mathcal{V}_{\text{mask}} \\ \boldsymbol{x}_i & v_i \notin \mathcal{V}_{\text{mask}}\end{cases}.
% \end{equation}

\begin{figure}[!t] % !htb
\begin{center}
\includegraphics[width=1.0\linewidth]{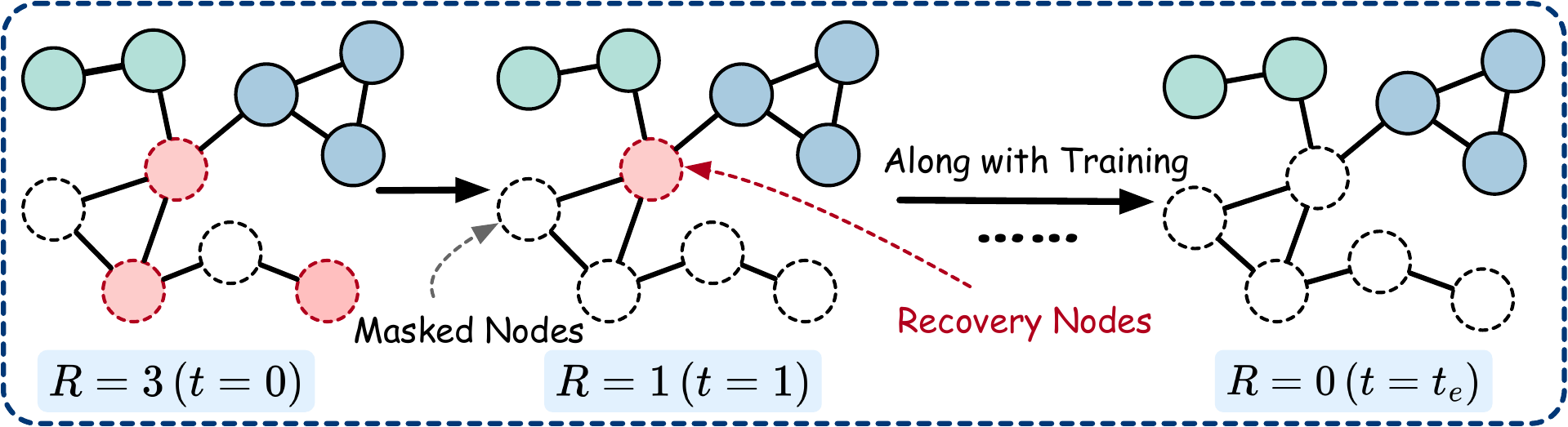}
\end{center}
\caption{Illustration of the dynamic node recovery mechanism employed in the CoFi masking. At each epoch, a certain number of nodes are randomly selected for recovery from those that have been masked. The number of recovered nodes (\textit{i.e.}, $R$) gradually decreases along with the training procedure.}
\label{fig:recover}
\end{figure}

\subsubsection{The Node Recovery Strategy}

The CoFi masking strategy, while effective in maintaining information consistency across hierarchical graph scales, poses certain challenges. Particularly, initiating masking at the coarsest scale may lead to the masking of entire subgraphs in the original-scale graph (in Figure~\ref{fig:masking} (b)). This extensive initial masking may hinder the model's effective reconstruction and learning from these subgraphs, particularly during the early stages of training. To address this issue, we integrate a recovery strategy into the CoFi masking approach. This strategy focuses on selectively unmasking certain nodes within the extensively masked subgraphs, enabling a more gradual and manageable learning process (in Figure~\ref{fig:recover}). Throughout the training process, we systematically reduce the number of nodes being recovered, denoted as $R^{(l,t)}$:
\begin{equation}
    R^{(l,t)} = n^{(l)}_{m} \cdot r_\text{re} \cdot \max \left\{ \left(1- \frac{t}{t_{\text{e}}}\right)^{\gamma}, 0 \right\}, \quad \gamma \geq 0,
    \label{eq:recover}
\end{equation}
where $n^{(l)}_{m}$ denotes the number of masked nodes at the $l$-th scale, $ r_\text{re}$ denotes the initial recovery ratio (set as 0.5), $t$ denotes the current epoch, $t_{\text{e}}$ denotes the end epoch for the recovery operation (set as a quarter of the total epochs), and $\gamma$ is the decay ratio (set as 1). This gradual decrease serves to progressively intensify the learning difficulty encountered by the models. Hence, the whole recovery strategy can be regarded as a warm-up mechanism for the model’s learning process. By initially presenting the model with a less challenging reconstruction task and gradually increasing the complexity, we experimentally demonstrate that the model can progressively adapt and improve its learning capabilities.

In summary, our approach to learning hierarchical information encompasses a sophisticated process of CoFi masking and recovery (CoFi-R). This comprehensive strategy enables effective learning of graph data at various scales, ensuring that the model progressively adapts and responds to increasing learning challenges.

\begin{figure*}[!t] % !htb
\begin{center}
\includegraphics[width=0.8\linewidth]{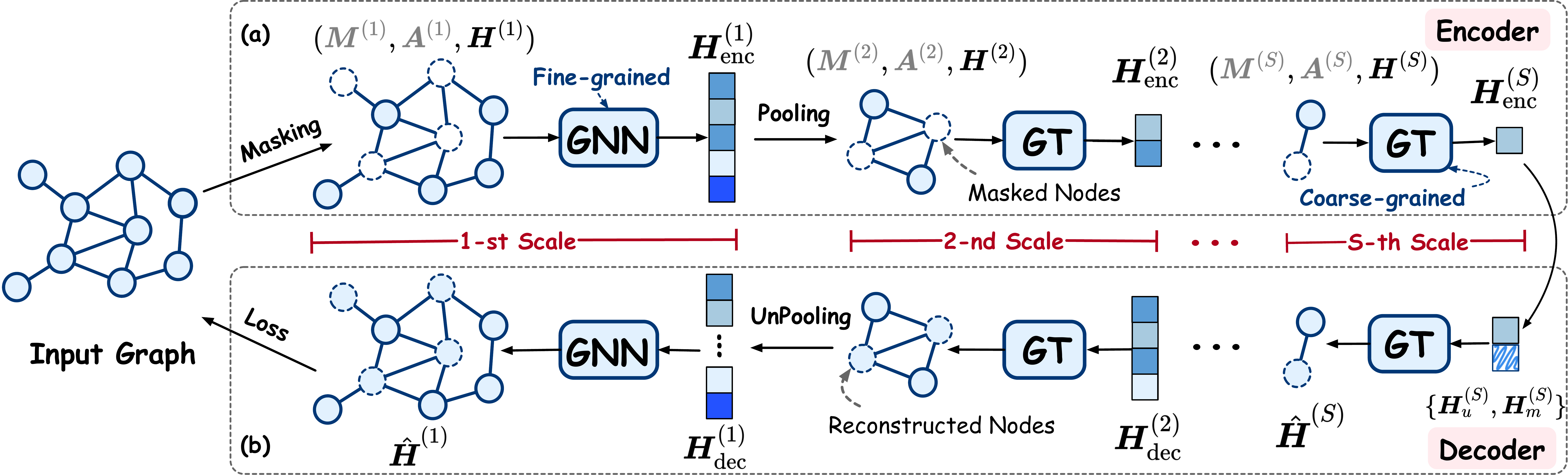}
\end{center}
\caption{Overview of the proposed model. \textbf{(a) Encoder:} The input graph with applied masking first undergoes fine-grained graph convolution (GNN). Following that, the graph undergoes a pooling process. Subsequently, a coarse-grained GT layer (GT) is applied to the coarse graph to facilitate the learning of high-level information. \textbf{(b) Decoder:} Upon encoding the unmasked nodes, we employ an unpooling strategy to progressively recover the original graph across different scales. For detailed technical descriptions of masking, pooling, and unpooling, please refer to Section~\ref{sec:propose-method}.   $({\color[HTML]{6D6D6D} \boldsymbol{M}^{(2)}}, {\color[HTML]{6D6D6D} \boldsymbol{A}^{(2)}})$  in {\color[HTML]{6D6D6D} grey} signifies that they undergo pre-processing before training.}
\label{fig:model}
\end{figure*}

\subsection{The Fine- and Coarse-Grained Encoder and Decoder}
\label{sec:encoder}

\subsubsection{The Encoder Process}  

Our model employs a multi-scale encoding strategy to process hierarchical graph data (shown in Figure~\ref{fig:model}), characterized by three distinct stages: 

\textbf{\text{ \large \ding{202}} Fine-Grained Graph Convolution Encoding:} The graphs with masked features $\boldsymbol{H}^{(l)} \in\mathbb{R}^{n \times d} $  are processed by a fine-grained graph convolution encoder, denoted as $\operatorname{GNN}_{\text{enc}}$ and referred to as GNN in Figure~\ref{fig:model}. The encoder comprises several GNN layers, such as Graph Isomorphism Networks (GIN)~\cite{gin}. These layers are designed to capture the local structures present within the masked graph. The process is formulated as below: 
\begin{equation}
\boldsymbol{H}_{\text{enc}}^{(l)}= \operatorname{GNN}_{\text{enc}}^{(l)}\left(\boldsymbol{A}^{(l)}, \boldsymbol{H}^{(l)},  \boldsymbol{M}^{(l)} \right) \in\mathbb{R}^{n^{(l)} \times d},
\label{eq:fine-gcn}
\end{equation}
where $\boldsymbol{H}_{\text{enc}}^{(l)}$ denotes the representations at the $l$-th coarse scale after the graph convolution, $n^{(l)}$ denotes the number of nodes in the graph at the $l$-th coarse scale, and  $\boldsymbol{H}^{(1)} = \boldsymbol{X}$.

\textbf{\text{ \large \ding{203}} Pooling Embeddings:} Following the fine-grained encoding, the output $\boldsymbol{H}_{\text{enc}}^{(l)}$ is subjected to a pooling module, denoted as $\operatorname{POOL}(\cdot)$. This module functions to project the embeddings to the next level in accordance with the predetermined assignment matrix $\boldsymbol{P}^{(l)}$. As a result of the pooling process, a condensed node embedding matrix for the graph at the $(l+1)$-th scale is generated:
\begin{equation}
\boldsymbol{H}^{(l+1)} = \mathrm{POOL}\left(\boldsymbol{H}_{\text{enc}}^{(l)} \right)  =  \left(\boldsymbol{P}^{(l)}\right)^{\top}  \boldsymbol{H}_{\text{enc}}^{(l)} \in \mathbb{R}^{n^{(l+1)} \times d}.
\label{eq:pooling}
\end{equation} 
These condensed representations present a high-level perspective of the original graph. Please note that during the pooling phase shown in Figure~\ref{fig:model}, pooling operations are performed solely on node embeddings, without necessitating the pooling of the graph structure.

\textbf{\text{ \large \ding{204}} Coarse-Grained Graph Transformer Encoding:} The coarse graph then becomes the input for the coarse-grained GT~\cite{graphgps,graphormer-v1,gapformer} encoder. This encoder is tailored to process the graph at a more abstract level, focusing on high-level information and global structures. Only the unmasked nodes in the coarse graphs are transmitted into the self-attention encoder: $\tilde{\boldsymbol{H}}^{(l+1)} =  \{\boldsymbol{H}^{(l+1)}[\mathrm{idx},:],  \boldsymbol{H}^{(l+1)}_{\text{pe}} \}$, where $\mathrm{idx}$ denotes the index of unmasked nodes, and $\boldsymbol{H}^{(l+1)}_{\text{pe}}$ is the positional encoding derived from the coarse adjacency matrix $\boldsymbol{A}^{(l+1)}$ through random-walk structural encoding~\cite{rwpe}.  The encoder process can be represented using the following formula:
\begin{equation}
\begin{aligned}
   \boldsymbol{U}^{(l+1)} = \frac{\left(\tilde{\boldsymbol{H}}^{(l+1)} \boldsymbol{W}_Q^{(l+1)}\right)\left(\tilde{\boldsymbol{H}}^{(l+1)} \boldsymbol{W}_K^{(l+1)}\right)^{\top}}{\sqrt{d}}; \\
\boldsymbol{H}_{\text{enc}}^{(l+1)}= \mathrm{Softmax}\left(\boldsymbol{U}^{(l+1)}\right) \tilde{\boldsymbol{H}}^{(l+1)} \boldsymbol{W}_V^{(l+1)},  
\end{aligned}
\label{eq:self-attention}
\end{equation}
where $\boldsymbol{H}_{\text{enc}}^{(l+1)} \in \mathbb{R}^{n^{(l+1)}_{\text{un}} \times d} $ denotes the representations of unmasked nodes at the $(l+1)$-th coarse scale after the GT layer. $\boldsymbol{W}_Q^{(l+1)} \in \mathbb{R}^{ d \times d} $,  $\boldsymbol{W}_K^{(l+1)} \in \mathbb{R}^{ d \times d} $,  and $\boldsymbol{W}_V^{(l+1)} \in \mathbb{R}^{ d \times d} $ are learnable parameters. And, $n^{(l+1)}_{\text{un}}$ denotes the number of unmasked nodes in the $(l+1)$-th scale. Please note that we omit layer normalization and skip connection in Eq.~(\ref{eq:self-attention}) due to space constraints. Then, the output $\boldsymbol{H}_{\text{enc}}^{(l+1)}$ is directed to the next set of pooling and coarse-grained GT encoding layers.

This sequential processing—from fine-grained graph convolutional encoding to pooling embeddings and coarse-grained graph transformer encoding—ensures that our model comprehensively learns the graph at a multi-scale level. 

\subsubsection{The Decoder Process} 
The decoding mechanism in our model plays a pivotal role in reconstructing the original graph from its unmasked encoded representations. 
% This process involves several key steps.

\textbf{\text{ \large \ding{202}} Coarse-grained Decoding:}
From the hierarchical encoder, we obtain the unmasked nodes' representations at all scales, represented as  $\{\boldsymbol{H}^{(l)}_{u} \}_{l=1}^S$, where $\boldsymbol{H}^{(l)}_{u} = \hat{\boldsymbol{H}}^{(l)}_{\text{dec}} $. Starting at the coarsest $S$-th scale, a shared learnable vector is assigned to each masked node, denoted as $\boldsymbol{H}^{(S)}_{m} \in \mathbb{R}^{n^{(S)}_{m} \times d^{(S)}}$. Then, these vectors are concatenated with the unmasked node representations, forming a combined representation  $\hat{\boldsymbol{H}}^{(S)}_{\text{dec}} = \{\boldsymbol{H}^{(S)}_{u},\boldsymbol{H}^{(S)}_{m} \}$. This concatenated representation $\hat{\boldsymbol{H}}^{(S)}_{\text{dec}}$ serves as the input for the decoder at the $S$-th scale:
 \begin{equation}
\hat{\boldsymbol{H}}^{(S)}=f_{\text{dec}}\left( \boldsymbol{A}^{(S)},  \hat{\boldsymbol{H}}^{(S)}_{\text{dec}}\right) \in \mathbb{R}^{n^{(S)} \times d},
\end{equation}
where $f_{\text{dec}}$ represents the decoder network, which is generally lighter than the corresponding encoder network.

\textbf{\text{ \large \ding{203}} Graph Unpooling to the Finer Scales:} The output $\hat{\boldsymbol{H}}^{(S)}$ undergoes an unpooling process, denoted as $\operatorname{UNPOOL}$. This unpooling step projects the representations back to the $(S-1)$-th scale:
\begin{equation}
\hat{\boldsymbol{H}}^{(S-1)}_{\text{dec}}=\operatorname{UNPOOL}\left(\hat{\boldsymbol{H}}^{(S)}\right) = \boldsymbol{P}^{(S-1)} \hat{\boldsymbol{H}}^{(S)} \in \mathbb{R}^{n^{(S-1)} \times d}.
\end{equation}
The above two processes are repeated iteratively, scaling from the coarsest scale to the original graph scale. Notably, the decoder choice, $f_{\text{dec}}(\cdot)$, can vary based on the scale. At the $S$-th scale, the decoder can be a GNN or GT layer. However, at other scales, the selected decoder is typically a GNN. This multi-scale approach is critical for capturing and reconstructing the nuanced details and broader structural patterns of the graph.
 
In summary, integrating fine-grained graph convolution with coarse-grained graph transformer modules in the encoder, along with a symmetrical unpooling process in the decoder defines the core functionality of our proposed model.  This architectural design enables comprehensive and multi-scale learning of the input graph, ensuring effective capture and processing of both detailed node-level information and broader graph structures.

\begin{table*}[!t]
\centering
\caption{Experimental results for \textbf{unsupervised representation learning} in graph classification. The results for baseline methods are sourced from prior studies. -- indicates the absence of corresponding results in the original paper. \textbf{Bold} or \underline{underline} indicates the best or second-best result, respectively, among self-supervised methods.}
% \vspace{-0.5em}
\label{tab:unspervise}
\renewcommand\arraystretch{1.0} % 行间距
\setlength\tabcolsep{3pt} % 列间距
\resizebox{0.99\textwidth}{!}{%
\begin{tabular}{@{}l|ccccccccc@{}}
\toprule
\multirow{1}{*}{} & \multirow{1}{*}{\textbf{PROTEINS}} & \multirow{1}{*}{\textbf{D\&D}} & \multirow{1}{*}{\textbf{NCI1}} 
& \multirow{1}{*}{\textbf{ENZYMES}} & \multirow{1}{*}{\textbf{MUTAGENCITY}} 
& \multirow{1}{*}{\textbf{IMDB-B}} &  \multirow{1}{*}{\textbf{IMDB-M}} & \multirow{1}{*}{\textbf{COLLAB}} & \multirow{1}{*}{\textbf{REDDIT-B}} \\ \midrule
%\midrule
%\rowcolor{Gray}
\multicolumn{10}{c}{\textit{Supervised Methods (2)}}\\
\midrule

% GCN~\cite{gcn}               & $79.68_{\pm 2.05}$    & $71.7_{\pm 4.7}$        & $73.4_{\pm 10.8}$      & $74.3_{\pm 4.6}$           & $71.92_{\pm 1.18}$     & $71.89_{\pm 0.33}$    & $69.50_{\pm 0.98}$  & $0.367_{\pm 0.011}$      \\
GIN~\cite{gin}            & $76.2_{\pm 2.8}$  & $70.8_{\pm 1.1}$  
& $82.7_{\pm 1.7}$       &$59.6_{\pm 4.5}$ &  $69.4_{\pm 1.2}$   & $75.1_{\pm 5.1}$                 & $52.3_{\pm 2.8}$     & $80.2_{\pm 1.9}$    & $92.4_{\pm 2.5}$    \\
DiffPool~\cite{diffpool}                & $75.1_{\pm 3.5}$     &  $77.6_{\pm 0.4}$   & $79.0_{\pm 1.8}$  &$59.5_{\pm 5.6}$ &  $77.6_{\pm 2.7}$ & $72.6_{\pm 3.9}$                 & $51.3_{\pm 0.7}$     & $78.9_{\pm 2.3}$    & $92.1_{\pm 2.6}$       \\
\midrule
\multicolumn{10}{c}{\textit{Contrastive-based Methods (10)}} \\
\midrule
Infograph~\cite{infoGraph:}               & $74.44_{\pm 0.31}$                              &  $72.85_{\pm 1.78}$                    & $76.20_{\pm 1.06}$     & -- & $72.32_{\pm 1.70}$        & $73.03_{\pm 0.87}$                         & $49.69_{\pm 0.53}$                       &$70.65_{\pm 1.13}$                       &$82.50_{\pm 1.42}$                                \\
GraphCL~\cite{graphCL}               & $74.39_{\pm 0.45}$                                 &   $78.62_{\pm 0.40}$                    & $77.87_{\pm 0.41}$            & $50.38_{\pm 0.34}$ & $71.77_{\pm 0.85}$      & $71.14_{\pm 0.44}$                      & $48.58_{\pm 0.67}$                       &$71.36_{\pm 1.15}$                       & $89.53_{\pm 0.84}$                      \\
AD-GCL~\cite{adgcl}  & $75.04_{\pm 0.48}$  & $75.73_{\pm 0.81}$  & $78.14_{\pm 0.62}$ & -- & -- & $71.49_{\pm 0.98}$  & $49.89_{\pm 0.66}$  & $74.89_{\pm 0.90}$  & $\textbf{92.35}_{\pm \textbf{0.42}}$ \\
JOAO~\cite{joao}        & $74.55_{\pm 0.41}$                                          &    $77.32_{\pm 0.54}$      & $78.07_{\pm 0.47}$   
& $43.67_{\pm 0.78}$ & $70.87_{\pm 0.77}$ &  $70.21_{\pm 3.08}$              &  $49.20_{\pm 0.77}$                      &     $69.50_{\pm 0.36}$                 &  $85.29_{\pm 1.35}$                                       \\ 
GCC~\cite{gcc}        & $69.5$   & $75.5$   &  $74.5$  & -- & $74.4$     &  $72.0$             &  $49.4$                      &    $78.9$                   &  $89.8$                                  \\ 
% MVGRL~\cite{mvgrl}               & --                                                 & --       & $89.70_{\pm 1.10}$              & $74.20_{\pm 0.70}$                       & $51.20_{\pm 0.50}$                       &--                       & $84.50_{\pm 0.60}$                    &$0.094_{\pm 0.008}$                       \\
InfoGCL~\cite{infogcl}               & --                                                 & --      &     $80.20_{\pm 0.60}$   & -- & --     &$75.10_{\pm 0.90}$                        & $51.40_{\pm 0.80}$                       &$80.00_{\pm 1.30}$                       & --                        \\
SimGRACE~\cite{simGrace}               & $75.35_{\pm 0.09}$                           &     ${77.44}_{\pm 1.11}$   & $79.12_{\pm 0.44}$  
 &     ${48.97}_{\pm 0.57}$   & $81.11_{\pm 0.33}$ 
 &$71.30_{\pm 0.77}$                        &$50.80_{\pm 0.68}$                        &$71.72_{\pm 0.82}$                       & $89.51_{\pm 0.89}$     \\
RGCL~\cite{rgcl} & $75.0_{\pm 0.4}$ & $78.9_{\pm 0.5}$ & $78.1_{\pm 1.0}$ & -- & -- & $71.9_{\pm 0.9}$ & -- & $71.0_{\pm 0.7}$ & $90.3_{\pm 0.6}$\\
CGKS~\cite{cgks}    &$76.0_{\pm 0.2}$  & -- & $79.1_{\pm 0.2}$ & -- & -- & -- & -- & $76.8_{\pm 0.1}$ & $91.7_{\pm 0.2}$ \\
DRGCL~\cite{DRGCL} &$75.2_{\pm 0.6}$ &$ 78.4_{\pm 0.7}$ &$78.7_{\pm 0.4}$
&$48.3_{\pm 0.6}$ &$72.7_{\pm 1.3}$ &$72.0_{\pm 0.5}$ &$48.8_{\pm 0.3}$ &$70.6_{\pm 0.8}$ &$90.8_{\pm 0.3}$ \\
\midrule
\multicolumn{10}{c}{\textit{Generative-based Methods (5)}} \\
\midrule
GraphMAE~\cite{graphmae}        & $75.30_{\pm 0.39}$                                    &      ${78.47}_{\pm 0.23}$             & $80.40_{\pm 0.30}$         &$49.17_{\pm 1.15}$  &$81.17_{\pm 0.16}$             &  $75.52_{\pm 0.66}$              &  ${51.63}_{\pm 0.52}$                     &   $80.32_{\pm 0.46}$                    &  $88.01_{\pm 0.19}$                            \\  
GraphMAE2~\cite{graphmae2} &$74.86_{\pm 0.34}$  &$78.66_{\pm 0.49}$ &$78.56_{\pm 0.26}$  &$49.13_{\pm 1.67}$ &$\textbf{81.70}_{\pm \textbf{0.30}}$  &$73.88_{\pm 0.53}$ &$51.80_{\pm 0.60}$  &$77.59_{\pm 0.62}$ &$76.84_{\pm 0.21}$    \\
S2GAE~\cite{s2gae}        & ${76.37_{\pm {0.43}}}$                                   &    $78.63_{\pm 0.81}$                & $80.80_{\pm 0.24}$   &--&--                   &  $75.76_{\pm 0.62}$              &  ${51.79_{\pm 0.36}}$                     &   ${81.02_{\pm 0.53}}$                    &  $87.83_{\pm 0.27}$                              \\
GCMAE~\cite{GCMAE} & -- & -- &  $81.42_{\pm 0.30}$ & -- & -- &   $\underline{75.78_{\pm 0.23}}$ &  $\underline{52.49_{\pm 0.45}}$ &  $81.32_{\pm 0.32}$ &  $\underline{91.75_{\pm 0.22}}$\\
AUG-MAE~\cite{augmae} &$75.83_{\pm 0.24}$  &$78.79_{\pm 0.40}$ &$78.48_{\pm 0.49}$  &$48.57_{\pm 0.86}$ &$76.86_{\pm 0.87}$  &$75.56_{\pm 0.61}$ &$51.80_{\pm 0.36}$  &$80.48_{\pm 0.50}$ &$87.98_{\pm 0.43}$   \\
\midrule
Hi-GMAE-G (ours)    & $\underline{{76.63}_{\pm 0.21}} $   &   $\underline{{80.03}_{\pm 0.45}}$ &  $\underline{82.00_{\pm 0.63}}$  &$\underline{50.53_{\pm 1.47}}$  &$81.18_{\pm 0.54}$      & ${\textbf{76.14}}_{\pm \textbf{0.19}}$      &                        ${ {51.93}}_{\pm 0.37}$ &   $ \underline{{82.16}_{\pm 0.17}}$   &  ${{88.81}}_{\pm 0.26}$                             \\
Hi-GMAE-F (ours)    & $\textbf{83.88}_{\pm \textbf{0.36}} $  &    $\textbf{83.68}_{\pm \textbf{0.39}}$  &  ${\textbf{82.21}_{\pm \textbf{0.36}}}$   &$\textbf{52.60}_{\pm \textbf{0.89}}$  &$\underline{81.45_{\pm 0.17}}$   & ${ 75.26_{\pm 0.63}}$      &                        ${ \textbf{52.64}_{\pm \textbf{0.18}}}$ &   ${ \textbf{83.60}_{\pm \textbf{0.84}}}$   &  ${88.22_{\pm 0.41}}$                            \\ \bottomrule
\end{tabular}
}
% \begin{minipage}{1.0\linewidth} \small % \footnotesize %\scriptsize %\scriptsize %\tiny
% \vspace{0.5em}
% Notations:  The results of baselines are from previous paper if avaliable.
% \end{minipage} 
% \vspace{-0.9em}
\end{table*}

\begin{table*}[!t]
\centering
\caption{Experimental results for \textbf{transfer learning} on molecular property prediction. The model is initially pre-trained on the ZINC15 dataset and subsequently fine-tuned on the above datasets. The reported metrics are ROC-AUC scores. The results for baseline methods are derived from previous studies. \textbf{Avg.} denotes the average performance.}
\label{tab:transfer}
\renewcommand\arraystretch{1.05} % 行间距
\setlength\tabcolsep{9pt} % 列间距
\resizebox{0.98\textwidth}{!}{%
\begin{tabular}{@{}l|cccccccc|c@{}}
\toprule
            & \textbf{BBBP}                                  & \textbf{Tox21}                                  & \textbf{ToxCast}                               & \textbf{SIDER}                                 & \textbf{ClinTox}                                  & \textbf{MUV}                          & \textbf{HIV}                                   & \textbf{BACE}                                                       & \textbf{Avg.}               \\ 
\midrule
\multicolumn{10}{c}{\textit{Supervised (no-pretrained) Methods (5)}} \\
\midrule
GCN~\cite{gcn} & $64.9 {\pm 3.0}$ & $74.9 {\pm 0.8}$ & $63.3 {\pm 0.9}$ & $60.0 {\pm 1.0}$ & $65.8 {\pm 4.5}$ & $73.2 {\pm 1.4}$ & $75.7 {\pm 1.1}$ & $73.6 {\pm 3.0}$ & 68.9 \\
GAT~\cite{gat} & $66.2 {\pm 2.6}$  & $75.4 {\pm 0.5}$ & $64.6 {\pm 0.6}$ & $60.9 {\pm 1.4}$ & $58.5 {\pm 3.6}$ & $66.6 {\pm 2.2}$ & $72.9 {\pm 1.8}$ & $69.7 {\pm 6.4}$ & 66.8 \\
SAGPool~\cite{sagpool} & $66.6 {\pm 2.2}$ & $72.2 {\pm 0.3}$  & $61.4 {\pm 1.2}$ & $60.5 {\pm 0.8}$ & $62.0 {\pm 2.8}$ & $70.4 {\pm 1.8}$  & $67.0 {\pm 1.2}$  & $73.4 {\pm 0.8}$ & 66.7 \\
GMT~\cite{gmt}  & $65.0 {\pm 3.9}$  & $72.3 {\pm 1.3}$  & $61.6 {\pm 0.9}$ & $56.8 {\pm 0.9}$  & $67.5 {\pm 5.7}$  & $67.2 {\pm 3.6}$  & $73.1 {\pm 1.8}$  & $74.5 {\pm 2.5}$ & 67.3\\
GraphGPS~\cite{graphgps}  & $69.2 {\pm 2.4}$  & $74.6 {\pm 0.8}$  & $63.1 {\pm 0.7}$  & $58.1 {\pm 1.0}$  & $64.4 {\pm 4.2}$  & $71.2 {\pm 2.0}$  & $76.0 {\pm 1.4}$  & $77.0 {\pm 2.1}$ & 69.2\\ 
Hi-GMAE-G (ours) & $70.4{\pm 2.4}$ & $74.6{\pm 0.8}$ & $63.1{\pm 0.7}$ & $58.1{\pm 1.0}$ & $64.4{\pm 4.2}$ & $71.2{\pm 2.0}$ & $76.0{\pm 1.4}$ & $77.0{\pm 2.1}$ & 69.4
\\ 
Hi-GMAE-F (ours) & $67.9{\pm 2.8}$                        & $74.1 {\pm 1.8}$                         & $62.2 {\pm 1.2}$                        & $58.3 {\pm 1.3}$                        & $56.4 {\pm 5.0}$                           & $62.1 {\pm 2.8}$               & $74.1 {\pm 1.5}$                        & $76.6 {\pm 3.6}$                        & 66.5    \\
\midrule
\multicolumn{10}{c}{\textit{Pretrained Methods (14)}} \\
\midrule
ContextPred~\cite{pretrain-gnn} & $64.3 {\pm 2.8}$                        & $75.7 {\pm 0.7}$             & $63.9 {\pm 0.6}$                        & $60.9 {\pm 0.6}$                        & $65.9 {\pm 3.8}$                           & $75.8 {\pm 1.7}$               & $77.3 {\pm 1.0}$                        & \multicolumn{1}{c|}{$79.6 {\pm 1.2}$}                        & 70.4               \\
AttrMasking~\cite{pretrain-gnn} & $64.3 {\pm 2.8}$                        & ${76.7} {\pm 0.4}$  & $64.2 {\pm 0.5}$ & $61.0 {\pm 0.7}$            & $71.8 {\pm 4.1}$                           & $74.7 {\pm 1.4}$               & $77.2 {\pm 1.1}$                        & \multicolumn{1}{c|}{$79.3 {\pm 1.6}$}                        & 71.1               \\
Infomax~\cite{pretrain-gnn}     & $68.8 {\pm 0.8}$                        & $75.3 {\pm 0.5}$                         & $62.7 {\pm 0.4}$                        & $58.4 {\pm 0.8}$                        & $69.9 {\pm 3.0}$                           & $75.3 {\pm 2.5}$               & $76.0 {\pm 0.7}$                        & \multicolumn{1}{c|}{$75.9 {\pm 1.6}$}                        & 70.3               \\
GraphCL~\cite{graphCL} & $69.7 {\pm 0.7}$                        & $73.9 {\pm 0.7}$                         & $62.4 {\pm 0.6}$                        & $60.5 {\pm 0.9}$                        & $76.0 {\pm 2.7}$                           & $69.8 {\pm 2.7}$               & $\textbf{78.5} {\pm \textbf{1.2}}$ & \multicolumn{1}{c|}{$75.4{ \pm 1.4}$}                        & 70.8               \\
AD-GCL~\cite{adgcl}    & $69.7 {\pm 0.5}$ & $74.9 {\pm 0.4}$ & $63.1 {\pm 0.7}$     & $61.5 {\pm 0.9}$ & $76.6 {\pm 2.8}$             & $72.2 {\pm 1.7}$   & $75.7 {\pm 1.1}$            & \multicolumn{1}{c|}{${75.6} {\pm {1.0}}$} & $71.2$ \\
JOAO~\cite{joao}        & $70.2 {\pm 1.0}$                        & $75.0 {\pm 0.3}$                         & $62.9 {\pm 0.5}$                        & $60.0 {\pm 0.8}$                        & $81.3 {\pm 2.5}$               & $71.7 {\pm 1.4}$               & $76.7 {\pm 1.2}$                        & \multicolumn{1}{c|}{$77.3 {\pm 0.5}$}                        & 71.9               \\
GraphLoG~\cite{graphlog}    & $\textbf{72.5} {\pm \textbf{0.8}}$ & $75.7 {\pm 0.5}$ & $63.5 {\pm 0.7}$                        & $61.2 {\pm 1.1}$ & $76.7 {\pm 3.3}$             & $76.0 {\pm 1.1}$   & $77.8 {\pm 0.8}$            & \multicolumn{1}{c|}{$\underline{{83.5} {\pm {1.2}}}$} & $73.4$ \\
SimGRACE~\cite{simGrace}    & $71.2 {\pm 1.1}$ & $74.4 {\pm 0.3}$ & $62.6 {\pm 0.7}$                        & $60.2 {\pm 0.9}$ & $75.5 {\pm 2.0}$             & $75.4 {\pm 1.3}$   & $75.0 {\pm 0.6}$            & \multicolumn{1}{c|}{${74.9} {\pm {2.0}}$} & $71.2$ \\
GraphMVP~\cite{graphmvp}    & $70.8 {\pm 0.5}$ & $74.9 {\pm 0.8}$ & $63.1 {\pm 0.5}$     & $60.2 {\pm 1.1}$ & $79.1 {\pm 2.8}$             & $77.7 {\pm 0.6}$   & $76.0 {\pm 0.1}$            & \multicolumn{1}{c|}{${79.3} {\pm {1.5}}$} & $72.6$ \\

RGCL~\cite{rgcl}    & $71.2 {\pm 0.9}$ & $75.3 {\pm 0.5}$ & $63.1 {\pm 0.3}$                        & $61.2 {\pm 0.6}$ & $85.0 {\pm 0.8}$             & $73.1 {\pm 1.2}$   & $77.3 {\pm 0.8}$            & \multicolumn{1}{c|}{${75.7} {\pm {1.3}}$} & $72.7$ \\

GraphMAE~\cite{graphmae}    & $72.0 {\pm 0.6}$            & $75.5 {\pm 0.6}$             & $64.1 {\pm 0.3}$            & $60.3 {\pm 1.1}$                        & $82.3 {\pm 1.2}$ & $76.3 {\pm 2.4}$ & $77.2 {\pm 1.0}$                        & $83.1 {\pm 0.9}$                                 & $73.8$ \\  
GraphMAE2~\cite{graphmae2}    & $71.6 {\pm 1.6}$            & $75.9 {\pm 0.8}$             & $\textbf{65.6} {\pm \textbf{0.7}}$           & $59.6 {\pm 0.6}$                        & $78.8 {\pm 3.0}$ & $\underline{78.5 {\pm 1.1}}$ & $76.1 {\pm 2.2}$                        & $81.0 {\pm 1.4}$ & $73.4$\\     
% S2GAE~\cite{s2gae}    & $67.6_ {\pm 2.0}$            & $69.6_ {\pm 1.0}$             & $58.7_ {\pm 0.8}$            & $55.4_ {\pm 1.3}$                        & $59.6_ {\pm 1.1}$ & $60.1_ {\pm 2.4}$ & $68.0_ {\pm 3.7}$                        & $68.6_ {\pm 2.1}$  
% & $63.5$ \\  
UnifiedMol~\cite{unifiedmol} & $70.4 {\pm 1.6}$ & $75.2 {\pm 0.4}$ & $63.3 {\pm 0.3}$ & \underline{$ 62.2 {\pm 1.1}$} & $81.3 {\pm 4.9}$ & $76.9 {\pm 1.5}$ & $77.5 {\pm 1.1}$ & $80.3 {\pm 2.2}$ & $73.4$ \\
% BET~\cite{bet} & $68.3 {\pm 1.2}$ & $76.1 {\pm 0.3}$ & $62.2 {\pm 0.4}$ & $ 63.4 {\pm 0.9}$ & $81.1 {\pm 6.3}$ & $76.2 {\pm 1.1}$ & $\textbf{78.5} {\pm \textbf{0.8}}$ & $80.9 {\pm 2.4}$ & $73.3$ \\
Mole-BERT~\cite{molebert}    & $71.9 {\pm 1.6}$            & $\underline{76.8 {\pm 0.5}}$             & $64.3 {\pm 0.2}$            & $\textbf{62.8} {\pm \textbf{1.1}}$                       & $78.9 {\pm 3.0}$ & $\textbf{78.6} {\pm \textbf{1.8}}$ & $78.2 {\pm 0.8}$                        & $80.8 {\pm 1.4}$                                 & $74.0$ \\\midrule \midrule
 Hi-GMAE-G (ours)   & $\textbf{72.5} {\pm \textbf{1.5}}$           & $76.3 {\pm 0.5}$             & ${64.5 {\pm 0.4}}$            & ${60.5 {\pm 0.7}}$                        & $\textbf{86.4} {\pm \textbf{1.0}}$ & $77.1 {\pm 1.8}$ & $77.3 {\pm 0.9}$                        & ${82.6 {\pm 1.0}}$                                  & $\underline{74.6}$ \\
 Hi-GMAE-F (ours)   & ${71.0 {\pm 1.1}}$            & $\textbf{76.9} {\pm \textbf{0.3}}$             & $\underline{65.3 {\pm 0.4}}$            & $62.0 {\pm 0.7}$                        & $\underline{85.6 {\pm {2.8}}}$ & $77.5 {\pm 0.8}$ & $\textbf{78.5} {\pm \textbf{0.6}}$                        & $\textbf{85.0} {\pm \textbf{0.6}}$                                 & $\textbf{75.2}$ \\
\bottomrule
\end{tabular}%
}
\end{table*}

\section{Experiment}
\label{sec:experiment}

% In this section, we conduct extensive experiments  on a variety of benchmark graph datasets to validate the effectiveness of the proposed method. 

\subsection{Unsupervised Representation Learning}

% \paragraph{Objective.} To assess the efficacy of the Hi-GMAE model, we first subject it to the unsupervised learning task, which highlights the model's feature extraction capabilities. This is necessary because high-quality and informative representations are crucial for various downstream graph tasks.

\paragraph{Settings.} \textbf{Datasets.} We utilize nine real-world datasets, comprising three in the bioinformatics domain (\textit{i.e.}, PROTEINS, D\&D, and ENZYMES), two in the molecular field (\textit{i.e.}, NCI1 and MUTAGENCITY), and four in the social sector (\textit{i.e.}, IMDB-B, IMDB-M, COLLAB, and REDDIT-B). \textbf{Baseline Models.} To demonstrate our proposed method's effectiveness, we compare Hi-GMAE with the following 17 baseline models: \textit{1) Two supervised models:} GIN~\cite{gin} and DiffPool~\cite{diffpool}; \textit{2) 10 contrastive models:} Infograph~\cite{infoGraph:}, GraphCL~\cite{graphCL}, AD-GCL~\cite{adgcl},  JOAO~\cite{joao}, GCC~\cite{gcc}, InfoGCL~\cite{infogcl}, SimGRACE~\cite{simGrace}, RGCL~\cite{rgcl}, CGKS~\cite{cgks},  and DRGCL~\cite{DRGCL}; \textit{3) Five generative models:} GraphMAE~\cite{graphmae}, GraphMAE2~\cite{graphmae2}, S2GAE~\cite{s2gae}, GCMAE~\cite{GCMAE}, and AUG-MAE~\cite{augmae}.    \textbf{Implementation Details.} In the pre-training
phase, our approach employs the same encoder type as
GraphMAE for the fine-grained layer.  In the evaluation phase, encoded graph-level representations are fed into a downstream LIBSVM~\cite{svm} classifier for graph classification, consistent with other baseline models. We report the mean accuracy across five different runs utilizing a 10-fold cross-validation strategy. Additional details on datasets, baseline models, and hyper-parameter settings are provided in the anonymous code repository.

\paragraph{Results.} The results are presented in detail in Table~\ref{tab:unspervise}, where Hi-GMAE-G refers to Hi-GMAE utilizing only GNN as the encoder, and Hi-GMAE-F utilizes the Fi-Co encoder. The following observations can be derived from these results: \textbf{1)} State-of-the-art Performance. Hi-GMAE outperforms existing self-supervised baselines on seven out of the nine datasets. Specifically, Hi-GMAE achieves improvements of 9.8\%, 6.1\%,  1.7\%, 4.4\%, and 2.8\% in accuracy on the PROTEINS, D\&D,  NCI1,  ENZYMES, and COLLAB datasets, respectively.  The competitive performance of both Hi-GMAE-F and Hi-GMAE-G emphasizes the importance of incorporating hierarchical information during pre-training, demonstrating the proposed Hi-GMAE approach's efficacy. \textbf{2)} Comparison with Supervised Methods. Remarkably, Hi-GMAE-F, though self-supervised, attains superior performance on six out of the nine datasets compared to supervised methods. This finding indicates that the representations learned by Hi-GMAE-F are high-quality and informative, aligning with the supervised learning benchmarks. \textbf{3)} Comparison with Generative Methods. Compared to GraphMAE~\cite{graphmae}, S2GAE~\cite{s2gae}, GCMAE~\cite{GCMAE}, and AUG-MAE~\cite{augmae} which are single-scale GMAE models, Hi-GMAE consistently outperforms them across all datasets. This result further supports our hypothesis that leveraging hierarchical structural information can significantly enhance the model's learning capabilities. \textbf{4)} Fi-Co Encoder vs. GNN Encoder. The superior performance of Hi-GMAE-F over Hi-GMAE-G across almost all the datasets (except for IMDB-B) highlights the advantage of integrating self-attention mechanisms to capture high-level information. The potential reason for the poor performance on the IMDB dataset is its small size, which may lead to the overfitting issue when applying the GT layer. This distinction illustrates the Fi-Co encoder's ability to capture broader graph information through the GT layer, while GNN encoders primarily focus on local information. In summary, these findings collectively affirm the Hi-GMAE method's core principles and its effectiveness in unsupervised graph representation learning.

\subsection{Transfer Learning}

% \paragraph{Objective.} The objective of assessing Hi-GMAE's performance in transfer learning task is to evaluate the model's ability to generalize knowledge acquired during pre-training to unseen datasets. 

\paragraph{Settings.} \textbf{Datasets.} During the initial pre-training phase, Hi-GMAE is trained on a dataset comprising two million unlabeled molecules obtained from the ZINC15~\cite{zinc15} dataset. Subsequently, the model undergoes fine-tuning on the eight classification benchmark datasets featured in the MoleculeNet dataset~\cite{moleculenet}. In our evaluation, we adopt a scaffold-split approach for dataset splitting, as outlined in~\cite{graphmae}. \textbf{Baseline Models.} To demonstrate our proposed method's effectiveness, we compare Hi-GMAE with the following 19 baseline models: \textit{1) Five no-pretrained models:} GCN~\cite{gcn}, GAT~\cite{gat}, SAGPool~\cite{sagpool}, GMT~\cite{gmt}, and GraphGPS~\cite{graphgps};   \textit{2) Three unsupervised models:} Infomax, AttrMasking, and ContextPred~\cite{pretrain-gnn}; \textit{3) Seven contrastive models:} GraphCL~\cite{graphCL}, AD-GCL~\cite{adgcl}, JOAO~\cite{joao},   GraphLOG~\cite{graphlog}, SimGRACE~\cite{simGrace}, GraphMVP~\cite{graphmvp}, and RGCL~\cite{rgcl}; \textit{4) Four generative models:} GraphMAE~\cite{graphmae}, GraphMAE2~\cite{graphmae2}, UnifiedMol~\cite{unifiedmol}, and Mole-BERT~\cite{molebert}. We report the baseline model results from previous papers according to research norms. \textbf{Implementation Details.} The experiments are conducted 10 times, and the mean and standard deviation of the ROC-AUC scores are reported. The details regarding datasets, baseline models, and hyper-parameters are comprehensively summarized in the anonymous code repository.

% This configuration complies with standard practices in the field and guarantees a fair comparison with baseline models.

\paragraph{Results.} The detailed results in Table~\ref{tab:transfer} offer insightful observations into Hi-GMAE's performance within the transfer learning context.
\textbf{1)} State-of-the-art Performance. The observed performance improvements with Hi-GMAE, utilizing both GNN and Fi-Co encoders, across a wide range of datasets underscore its efficiency and adaptability. Specifically, Hi-GMAE-F and Hi-GMAE-G achieve a 1.6\% and 0.8\% improvement in average performance metrics, respectively. Additionally, each method individually achieves top-tier performance on several datasets. These results validate Hi-GMAE's ability to effectively generalize learned representations across diverse datasets. This makes it a potentially valuable tool for applications where labeled data is scarce or models need to rapidly adapt to new tasks. The results reinforce the significance of hierarchical structure learning in graphs. By capturing multi-scale information, Hi-GMAE can create representations that are more nuanced and informative, which is particularly beneficial for handling complex graph-based tasks. \textbf{2)} The Superiority of the Fi-Co Encoder. The results reveal the superior performance of Hi-GMAE-F over Hi-GMAE-G in  average metrics and across six out of the eight datasets. This suggests that the Fi-Co encoder's ability to capture high-level graph information offers a significant advantage over the GNN encoders used in other GMAE methods (\textit{e.g.}, GraphMAE~\cite{graphmae} and Mole-BERT~\cite{molebert}), aligning with previous observations in unsupervised learning. \textbf{3)} Comparison with Supervised Methods. We further conduct comprehensive comparisons of our Hi-GMAE model against a range of unpretrained supervised models, including graph convolutional models such as GCN~\cite{gcn} and GAT~\cite{gat}, graph transformer models like GraphGPS~\cite{graphgps}, and graph pooling models such as SAGPool~\cite{sagpool} and GMT~\cite{gmt}. The experimental results consistently show that pre-trained models, such as Hi-GMAE, significantly outperform both their unpretrained counterparts and SOTA supervised models in terms of accuracy. This consistent improvement across diverse model types and datasets highlights the significant benefits of pre-training.

\begin{table}[!t]
\centering
\caption{\textbf{Ablation study} of Hi-GMAE components. ``w/o CoFi'' means that we adopt random masking at each level. ``w/o Embedding Aggregation'' means that only the representation from the first scale is used as the final representation during inference.}
\label{tab:ablation}
\renewcommand\arraystretch{1.25} % 行间距
\setlength\tabcolsep{3pt} % 列间距
\resizebox{0.48\textwidth}{!}{%
\begin{tabular}{l|ccc}
\hline
                  & PROTEINS          & D\&D              & REDDIT-B        \\ \hline
\textbf{Hi-GMAE }          & $\textbf{83.88}_ {\pm \textbf{0.36}}$  & $\textbf{83.68}_ {\pm \textbf{0.39}}$  & $\textbf{88.22}_ {\pm \textbf{0.41}}$  \\ \hline
\quad w/o CoFi Masking & $81.44_ {\pm 0.25}$  & $82.11_ {\pm 0.51}$  & $86.50_ {\pm 1.55}$  \\
\quad w/o Node Recovery      & \underline{$83.39_ {\pm  0.34}$} & \underline{$82.24_ {\pm 0.40}$}  & \underline{$87.07_ {\pm 1.94}$}  \\
\quad w/o Embedding Aggregation       & $82.51_ {\pm  0.89}$ & $81.17_ {\pm  0.48}$ & $79.40_ {\pm  2.80}$ \\
% \quad w/o Skip-Connection    & $82.44_ {\pm  0.88}$ & $81.83_ {\pm  0.45}$ & $76.81_ {\pm  1.46}$ \\ 
\hline
\end{tabular}%
}
\end{table}

\begin{figure}[!t] % !htb
\begin{center}
\includegraphics[width=0.97\linewidth]{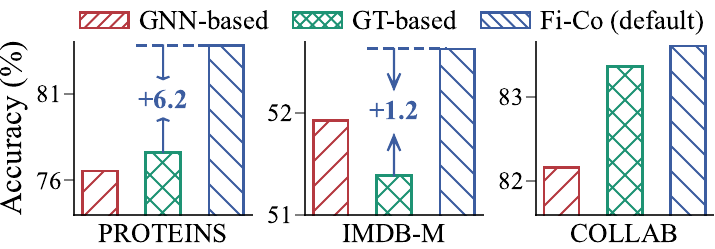}
\end{center}
\caption{Comparison of different encoder architectures on three datasets. Fi-Co denotes our proposed fine- and coarse-grained architecture. GNN-based/GT-based characterizes the approach of solely employing GNN/GT as the encoder at every scale.}
\label{fig:ablation-encoder}
\end{figure}

\subsection{Further Discussions and Analysis}
\label{sec:ablation study}

\textbf{The Impact of the Hi-GMAE Components.}
A detailed study is conducted to evaluate the impact of different components within Hi-GMAE. It is important to note that, except for the components under analysis, all other aspects of the model remain consistent with the comprehensive Hi-GMAE. The results in Table~\ref{tab:ablation} and Figure~\ref{fig:ablation-encoder}, as outlined below, provide valuable insights into the significance of each component: \textbf{1)} The CoFi Masking Strategy. The observed performance reduction in the absence of the CoFi masking strategy (\textit{i.e.}, 2.9\% on PROTEINS, 2.6\% on D\&D, and 1.9\% on REDDIT-B) underscores its importance.  This strategy significantly facilitates capturing the hierarchical graph structure by ensuring the masked information's consistency across all scales.  \textbf{2)} The Node Recovery. The positive impact of the node recovery component on Hi-GMAE's learning process supports the premise that introducing complexity in a phased manner can significantly enhance model learning. This staged complexity approach allows the model to gradually adapt and refine its understanding of the graph's structure, leading to improved performance. \textbf{3)} Multi-Scale Representation Aggregation.  Aggregating representations from different scales is especially beneficial for complex datasets like REDDIT-B. This process enables the model to integrate information from various granularity levels, such as from finer details to broader structural patterns, thereby enriching the final graph representation with comprehensive insights. \textbf{4)} The Fi-Co Encoder. Compared to the two variants that solely employ GNN or GT architecture as the encoder, our Fi-Co encoder demonstrates superior performance across all the evaluated datasets. This demonstrates that our Fi-Co model is effective at capturing multi-level graph information. Specifically, compared to GNN-based methods, Fi-Co incorporates a graph transformer, thus inheriting advanced expressive capabilities~\cite{wl-test} enabled by the design of position encodings and structure encodings in the graph transformer~\cite{rwpe,graphormer-v1}. In contrast to GT-based model, Fi-Co captures more local information while significantly reducing computational complexity. In summary, the above findings validate the individual contributions of each component and emphasize their synergy in capturing hierarchical graph data.  

% Due to space constraints, additional parameter analyses are provided in the Appendix.

% \textbf{4) Skip Connections:} The inclusion of skip connections within the Hi-GMAE architecture offers a notable boost in accuracy by facilitating the flow of complementary information to the decoder. 

\section{Conclusion} 
This study proposes the Hi-GMAE model, a novel multi-scale generative graph pre-training method, featuring several key elements: the generation of multi-level coarse graphs, a CoFi masking strategy with recovery, and a Fi-Co encoder. Through comprehensive experiments covering two distinct graph learning tasks, it is shown that Hi-GMAE excels at capturing hierarchical structural information and significantly outperforms existing self-supervised pre-training methods. These results emphasize the importance of considering the hierarchical structural information inherent in graphs during pre-training. Despite its competitive performance, Hi-GMAE still has room for improvement in several areas, including: 1) improving pooling methods to more effectively exploit hierarchical structural information,  and 2) expanding Hi-GMAE's applicability to a wide range of graph tasks (\textit{e.g.}, node classification and link prediction).

\begin{acks}
This work was supported in part by the Natural Science Foundation of China (No. 62476203, 62502065), Key Project of Traditional Chinese Medicine Joint Fund of Hubei Provincial Natural Science Foundation (No.2025AFD47), Hubei Province Science and Technology Innovation Plan Project (No.2025BCB035), the Guangdong Provincial Natural Science Foundation General Project (No. 2025A1515012155), the Shenzhen Natural Science Foundation Project (No. JCYJ20250604122534006, JCYJ20230807090211021), and partially supported by the ARC Projects with Nos. DP260100944 and FT250100159.
\end{acks}

\bibliographystyle{ACM-Reference-Format}
\bibliography{sample-base}

% \clearpage

\appendix

\section{Additional Experiments}

\subsection{Transfer Learning for Graph Regression}

\paragraph{Settings} \textbf{Datasets.} Our study involves two types of regression tasks: molecular property prediction and quantum chemistry property prediction. For molecular property prediction, we initially pre-train the Hi-GMAE model on the GEOM dataset~\cite{geom} and subsequently fine-tune it using two specific datasets from MoleculeNet~\cite{moleculenet}, namely Malaria and CEP. In the realm of quantum chemistry property prediction, we train Hi-GMAE using the ZINC15 dataset~\cite{zinc15} and fine-tune it on three additional datasets: QM7~\cite{qm7}, QM8~\cite{qm8}, and QM9~\cite{qm9}. \textbf{Baseline Models.} To validate the efficacy of our proposed Hi-GMAE model, we compare its performance against seven benchmark models, categorized as follows: \textit{1) Two unsupervised models:} AttrMasking~\cite{pretrain-gnn} and ContextPred~\cite{pretrain-gnn}; \textit{2) Three contrastive models:} GraphCL~\cite{graphCL}, JOAO~\cite{joao}, and GraphMVP~\cite{graphmvp}; \textit{3) Two generative models:} GraphMAE~\cite{graphmae} and Mole-BERT~\cite{molebert}. We adhere to research norms by reporting results from previous studies for these baseline models. \textbf{Implementation Details.} We employ a scaffold-split strategy for dataset division as detailed in GraphMVP~\cite{graphmvp}. The experiments are conducted three times to ensure reliability, and we report both the mean and standard deviation of the  Root Mean Square Error (RMSE) and Mean Absolute Error (MAE) metrics.

\paragraph{Results} We present the graph regression results under transfer learning settings in Table~\ref{tab:regression}. The data clearly shows that Hi-GMAE achieved state-of-the-art results on the majority of the datasets evaluated, outperforming existing methods in three out of the five datasets. Specifically, Hi-GMAE-F demonstrated significant improvements, achieving a 10.3\% increase in accuracy on the CEP dataset and a 5.1\% increase on the QM7 dataset. These results not only validate Hi-GMAE’s superior performance but also substantiate its ability to effectively generalize learned representations across diverse datasets. Such capabilities underscore the versatility of Hi-GMAE in handling various graph-based learning tasks under transfer learning scenarios.

\begin{table*}[!h]
\centering
\caption{Experimental results for \textbf{Graph regression} task on molecular property prediction and quantum chemistry property prediction. The results for baseline methods are sourced from prior studies. -- indicates the absence of corresponding results in the original paper. \textbf{Bold} or \underline{underline} indicates the best or second-best result, respectively.}
\label{tab:regression}
\renewcommand\arraystretch{1.2} % 行间距
\setlength\tabcolsep{6pt} % 列间距
\resizebox{0.9\textwidth}{!}{%
\begin{tabular}{@{}l|ccccc@{}}
\toprule
            & \textbf{Malaria (RMSE $\downarrow$)}           & \textbf{CEP (RMSE $\downarrow$)}               & \textbf{QM7 (MAE $\downarrow$)}                        & \textbf{QM8 (MAE $\downarrow$)}                & \textbf{QM9 (MAE $\downarrow$)}             \\ \midrule
ContextPred~\cite{pretrain-gnn} & $1.119_{\pm 0.014}$ & $1.243_{\pm 0.025}$ & $88.3_{\pm 0.5}$             & $0.0196_{\pm 0.0002}$ & $9.14_{\pm 0.40}$  \\
AttrMask~\cite{pretrain-gnn}    & $1.101_{\pm 0.015}$ & $1.256_{\pm 0.000}$ & $109.5_{\pm 7.6}$            & $0.0194_{\pm 0.0001}$ & $11.74_{\pm 0.73}$ \\
GraphCL~\cite{graphCL}     & -                 & -                 & $80.4_{\pm 3.3}$             & $0.0200_{\pm 0.0004}$ & $5.76_{\pm 0.37}$  \\
JOAO~\cite{joao}        & $1.145_{\pm 0.010}$ & $1.293_{\pm 0.003}$ & $84.3_{\pm 2.1}$             & $0.0206_{\pm 0.0001}$ & $10.65_{\pm 0.79}$ \\
GraphMVP~\cite{graphmvp}    & $1.106_{\pm 0.013}$ & \underline{$1.228_{\pm 0.001}$} & -                          & -                   & -                \\
GraphMAE~\cite{graphmae}    & $1.098_{\pm 0.012}$ & $1.263_{\pm 0.010}$ & \underline{$78.4_{\pm 2.6}$} & $\textbf{0.0190}_{\pm \textbf{0.0003}}$ & $5.84_{\pm 0.16}$  \\
Mole-BERT~\cite{molebert}   & $\textbf{1.074}_{\pm \textbf{0.009}}$ & $1.232_{\pm 0.009}$ & $79.8_{\pm 2.6}$             & $\textbf{0.0190}_{\pm \textbf{0.0003}}$ & \underline{$5.75_{\pm 0.16}$}  \\ \midrule
Hi-GMAE-F (ours)    & \underline{$1.091_{\pm 0.015}$} & $\textbf{1.101}_{\pm \textbf{0.012}}$ & $\textbf{74.4}_{\pm \textbf{1.9}}$             & $0.0194_{\pm 0.0003}$ & $\textbf{5.47}_{\pm \textbf{0.12}}$  \\ \bottomrule
\end{tabular}%
}
\end{table*}

\section{Complexity Analysis}
In this section, we explore the computational complexity of Hi-GMAE, particularly in comparison with GraphMAE~\cite{graphmae}. As outlined above, the generation of coarse graphs and the corresponding masks are executed as pre-processing steps. This preliminary processing prepares the graph data for efficient handling in the subsequent stages of the Hi-GMAE model. The primary training distinction between Hi-GMAE and GraphMAE is the inclusion of pooling and unpooling operations. At the $l$-th scale, Hi-GMAE requires $\mathcal{O}\left(r_{p}^{(l-1)} n \cdot r_{p}^{(l)} n\right)$. With a constant pooling ratio $r_p$ across all scales, the complexity for all $S$ scales is simplified to  $\mathcal{O}\left(r_{p} n^2+r_{p}^3 n^2+\cdots+r_{p}^{2 S-1} n^2\right)$. Considering that $r_p < 1$, the terms $\mathcal{O}\left(r_{p}^3 +\cdots+r_{p}^{2 S-1} \right)$ are significantly smaller than 1, reducing the overall complexity of pooling/unpooling process to approximately $\mathcal{O}\left(r_{p} n^2 \right)$. 

Furthermore, the complexity of Hi-GMAE's coarse-grained encoding module (\textit{i.e.}, self-attention) is quadratic \textit{w.r.t.} the number of input nodes. This complexity exceeds that of traditional GCN-based encoders. However, a crucial optimization within Hi-GMAE ensures that only the coarse graph's unmasked nodes are processed by the coarse module, thereby adjusting its complexity to $\mathcal{O}(r_{m}^{2}  r_{p}^{2} n^2)$, where $r_{m}$ represents the mask ratio. Notably, when appropriate ratios are selected, so that $r_{p} n \ll n$ and $r_{m}^{2}  r_{p}^{2} n \ll n$ (\textit{e.g.}, $r_{m} = 0.2$,  $r_{p} = 0.2$, and $n=40$), the overall complexity becomes almost linear to $n$, making Hi-GMAE a viable and efficient GMAE model.

\section{Efficiency Analysis}

Table~\ref{tab:time_cost} outlines the GPU memory usage and training time on the NCI1 and IMDB-B datasets, with a comparative analysis between Hi-GMAE, GraphMAE~\cite{graphmae}, and AUG-MAE~\cite{augmae}. The key observations from this comparison are as follows: \textbf{1) Memory Usage:} Hi-GMAE displays a slightly higher memory consumption compared to GraphMAE. This increase is primarily due to the integration of GT layers as encoders within Hi-GMAE, which are inherently more memory-intensive than conventional GNNs. This architectural choice enhances performance but necessitates greater memory overhead. \textbf{2) Time Consumption:}  Hi-GMAE has a slightly higher training time compared to GraphMAE, but it is significantly lower than that of AUG-MAE. During training, Hi-GMAE processes multiple graphs in batches, which requires the aggregation of these coarse-grained adjacency matrices into a single batch-style adjacency matrix. This step introduces additional computational demands and extends the overall time required for training. Optimizing this aggregation operation will be a primary focus in our future research efforts to enhance efficiency and reduce time consumption.

\begin{table}[!h]
\centering
\caption{Detailed Time \& Memory of Hi-GMAE and other baseline methods on two datasets.}
\label{tab:time_cost}
\renewcommand\arraystretch{1.2} % 行间距
\setlength\tabcolsep{4pt} % 列间距
\resizebox{0.48\textwidth}{!}{%
\begin{tabular}{@{}c|cccccc@{}}
\toprule
 & \multicolumn{2}{c}{\textbf{NCI1}} & \multicolumn{2}{c}{\textbf{IMDB-B}} \\ \cmidrule(lr){2-3} \cmidrule(lr){4-5}  
                & Time (s) & Memory (MB) & Time (s) & Memory (MB) \\ \midrule
GraphMAE~\cite{graphmae}          & 1,510 & 576     & 53   & 649      \\
AUG-MAE~\cite{augmae}         & 1,762 & 1,928  & 75   & 1,656     \\ \midrule
Hi-GMAE (ours)   & 2,652 & 682  & 70   & 677        \\ \bottomrule
\end{tabular}%
}
\end{table}

\section{Ablation Study}
\label{sec:hier_study}

\begin{figure}[!h] % !htb
\begin{center}
\includegraphics[width=0.97\linewidth]{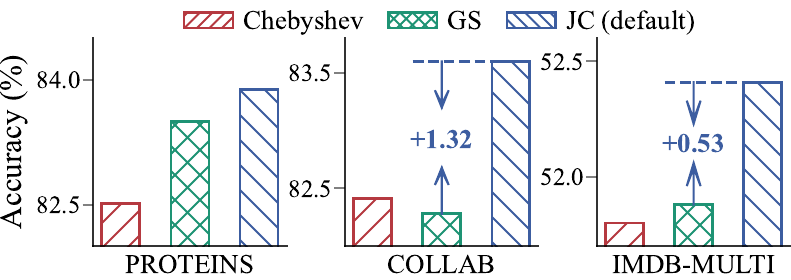}
\end{center}
\caption{The results of Hi-GMAE with different pooling methods across three datasets. Adopted three pooling methods are derived from ANS-GT.}
\label{fig:coarse_method}
\end{figure}

\textbf{The Impact of the Pooling Methods.}  We assess three graph pooling algorithms (\textit{i.e.}, $\mathrm{POOL(\cdot)}$ in Eq.~(\ref{eq:pooling})), namely Chebyshev, Algebraic JC (JC), and Gauss-Seidel (GS). Our evaluations, conducted on the PROTEINS, COLLAB, and IMDB-M datasets and illustrated in Figure~\ref{fig:coarse_method}, demonstrate that the JC method consistently surpasses both the Chebyshev and GS methods across all datasets. Notably, the JC method achieves a performance advantage of 0.5\%, 1.4\%, and 1.5\% over the next best method on the PROTEINS, COLLAB, and IMDB-M datasets, respectively. Based on these results, we have designated the JC method as the standard graph pooling technique for the Hi-GMAE model.

\section{Parameter Analysis}
\label{sec:parameter_ana}

\begin{figure}[!h] % !htb
\begin{center}
\includegraphics[width=0.49\textwidth]{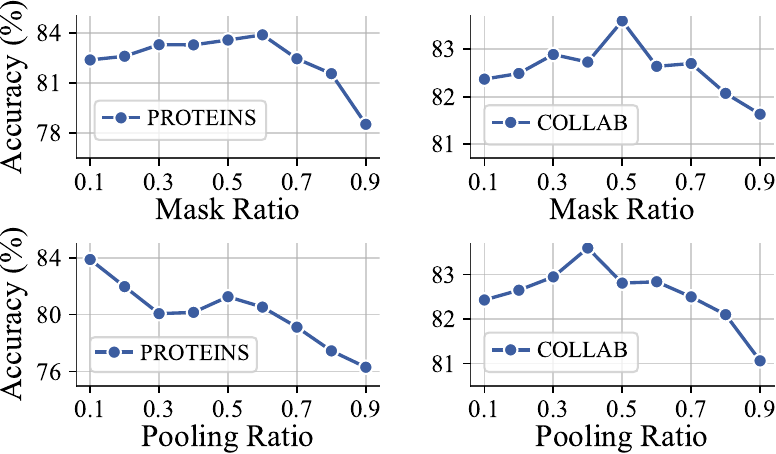}
\end{center}
\caption{Parameter analysis of the mask ratio and pooling ratio on two datasets.}
\label{fig:parameter-analysis-2}
\end{figure}

\textbf{Mask Ratio and Pooling Ratio.}
In this section, we delve into a comprehensive analysis of how  mask and pooling ratio influence the performance of the Hi-GMAE model. The insights derived from Figure~\ref{fig:parameter-analysis-2} reveal that the optimal mask ratio differs across datasets. We observe an initial improvement in Hi-GMAE's accuracy as the mask ratio increases, suggesting that masking a certain amount of information fosters enhanced learning capabilities by compelling the model to generalize more effectively from observed to masked data. However, this trend reverses once the mask ratio exceeds a specific threshold; beyond this point, accuracy diminishes, indicating that excessive masking may obstruct the model's capacity to accurately infer missing information, thus degrading overall performance. We recommend a mask ratio between 0.4 and 0.6 for deploying Hi-GMAE across various datasets. This range effectively balances the need for a challenging learning environment with the retention of sufficient information for accurate reconstruction and inference. Additionally, the pooling ratio (\textit{i.e.}, $r_p$), a critical hyperparameter in Hi-GMAE, determines the degree of graph size reduction during the coarsening process. According to the results depicted in Figure~\ref{fig:parameter-analysis-2}, we find that: \textbf{1)} The PROTEINS dataset achieves its best performance at a pooling ratio of 0.1, while the COLLAB dataset performs optimally at a ratio of 0.4. Beyond these optimal values, performance gradually declines. \textbf{2)} For both the PROTEINS and COLLAB datasets, the results indicate that lower pooling ratios, implying less aggressive graph reduction, generally lead to superior performance. Consequently, for unsupervised tasks, we suggest maintaining the pooling ratio within the range of 0.1 to 0.5, aiming to optimize the balance between data reduction and the preservation of essential structural details.

\begin{figure}[!h] % !htb
\begin{center}
\includegraphics[width=0.48\textwidth]{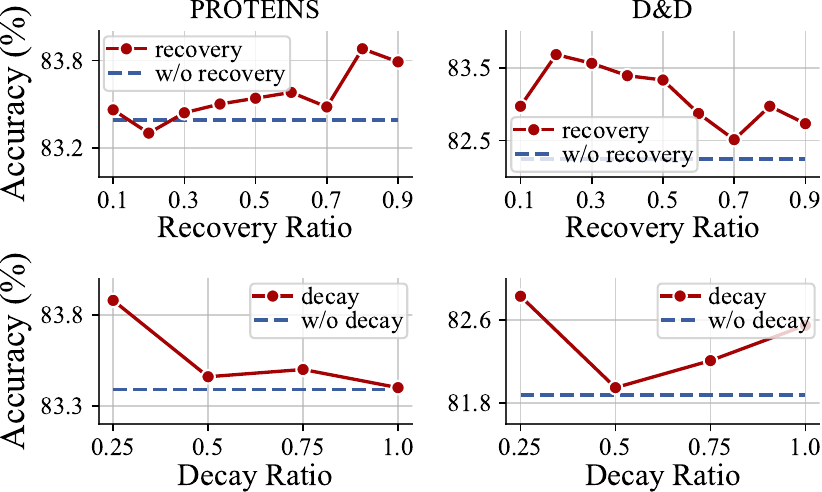}
\end{center}
\caption{Parameter analysis of the recovery ratio and decay ratio on two datasets (PROTEINS and D\&D).}
\label{fig:parameter-analysis}
\end{figure}

\noindent \textbf{Recovery Ratio and Decay Ratio.} The recovery ratio (\textit{i.e.}, $r_\text{re}$ in Eq.~(\ref{eq:recover})) determines the initial restoration proportion in Hi-GMAE. We explore a range of recovery ratios within $\{0.1, 0.2, ..., 0.9\}$. From the detailed statistics in Figure~\ref{fig:parameter-analysis}, we can infer that: \textbf{1)} Utilizing a low recovery ratio results in an insufficient restoration of nodes in the masked subgraphs, which can significantly limit the model's capability to learn necessary information from these critical areas.  \textbf{2)} The CoFi masking strategy with recovery strategy consistently outperforms the method without recovery across different recovery ratios. This consistency reaffirms our initial finding and underscores the effectiveness of the node recovery strategy in enhancing the model's accuracy.  Besides, the decay ratio (\textit{i.e.}, $\gamma$ in Eq.~(\ref{eq:recover})) is searched within $\{0.25, 0.5, 1, 2, 4\}$, which includes both linear and non-linear decay options. Figure~\ref{fig:parameter-analysis} indicates that a linear decay (\textit{i.e.}, $\gamma =1.0$) outperforms other configurations. Based on these observations, for the sake of simplicity, we set the decay ratio as 1.0 in all unsupervised tasks.

\noindent \textbf{The Number of Pooling Scale.} The varying numbers of scales (\textit{i.e.}, $S$) enable Hi-GMAE to capture diverse hierarchical information in the graph. Analysis of the results presented in Table~\ref{tab:coarse_layer} reveals two key insights: \textbf{1)} The integration of hierarchical information through multiple coarsening layers leads to a marked enhancement in the model's experimental performance. Specifically, compared to the baseline performance of the single-scale GraphMAE~\cite{graphmae}, there is a notable increase in performance metrics by 11.4\%, 4.1\%, and 1.5\%. This improvement underscores the value of incorporating multi-scale graph representations to achieve a deeper understanding and more accurate modeling of graph structures. \textbf{2)} Across the datasets examined, the optimal configuration for the Hi-GMAE model involves using 2 to 3 pooling scales. For instance, considering the NCI1 dataset, an increase in the number of scales leads to a reduction in the average number of nodes per graph (from 29.8 to 24.3 to 20.1). This trend results in a decreased variance in node counts before and after the coarsening process. This pattern may explain why the model demonstrates peak performance at 2-3 scales; beyond this range, the additional reduction in nodes yields diminishing returns, providing minimal additional value for extracting meaningful insights from the data.

\begin{table}[!h]
    \centering
    \caption{The impact of the pooling scale on three datasets.}
    \label{tab:coarse_layer}
    \renewcommand\arraystretch{1.2} % 行间距
    \setlength\tabcolsep{8pt} % 列间距
    \resizebox{0.48\textwidth}{!}{%
    \begin{tabular}{@{}l|ccc@{\hspace{0.3em}}}
    \toprule
     & \multicolumn{1}{c}{\textbf{PROTEINS}} & \multicolumn{1}{c}{\textbf{D\&D}} &  \multicolumn{1}{c}{\textbf{IMDB-M}}  \\
    \midrule
        GraphMAE~\cite{graphmae} & $75.30_{\pm 0.39}$  & $78.47_{\pm 0.23}$  & \underline{$51.63_{\pm 0.52}$}\\ \midrule
        Hi-GMAE (2-scale) & $\textbf{83.88}_{\pm \textbf{0.36}}$ & $82.36_{\pm 0.81}$ &  $50.87_{\pm 0.78}$ \\
        Hi-GMAE (3-scale) & \underline{$76.97_{\pm 0.49}$} & $\textbf{83.68}_{\pm \textbf{0.39}}$  & $\textbf{52.64}_{\pm \textbf{0.18}}$ \\
        Hi-GMAE (4-scale) & $74.48_{\pm 1.20}$ & \underline{$83.07_{\pm 1.13}$}  & $51.53_{\pm 0.92}$ \\
    \bottomrule
    \end{tabular}
     }
\end{table}

\section{The Visualization of Learned Representations} 
\begin{figure}[!h] % !htb
\begin{center}
\includegraphics[width=1.0\linewidth]{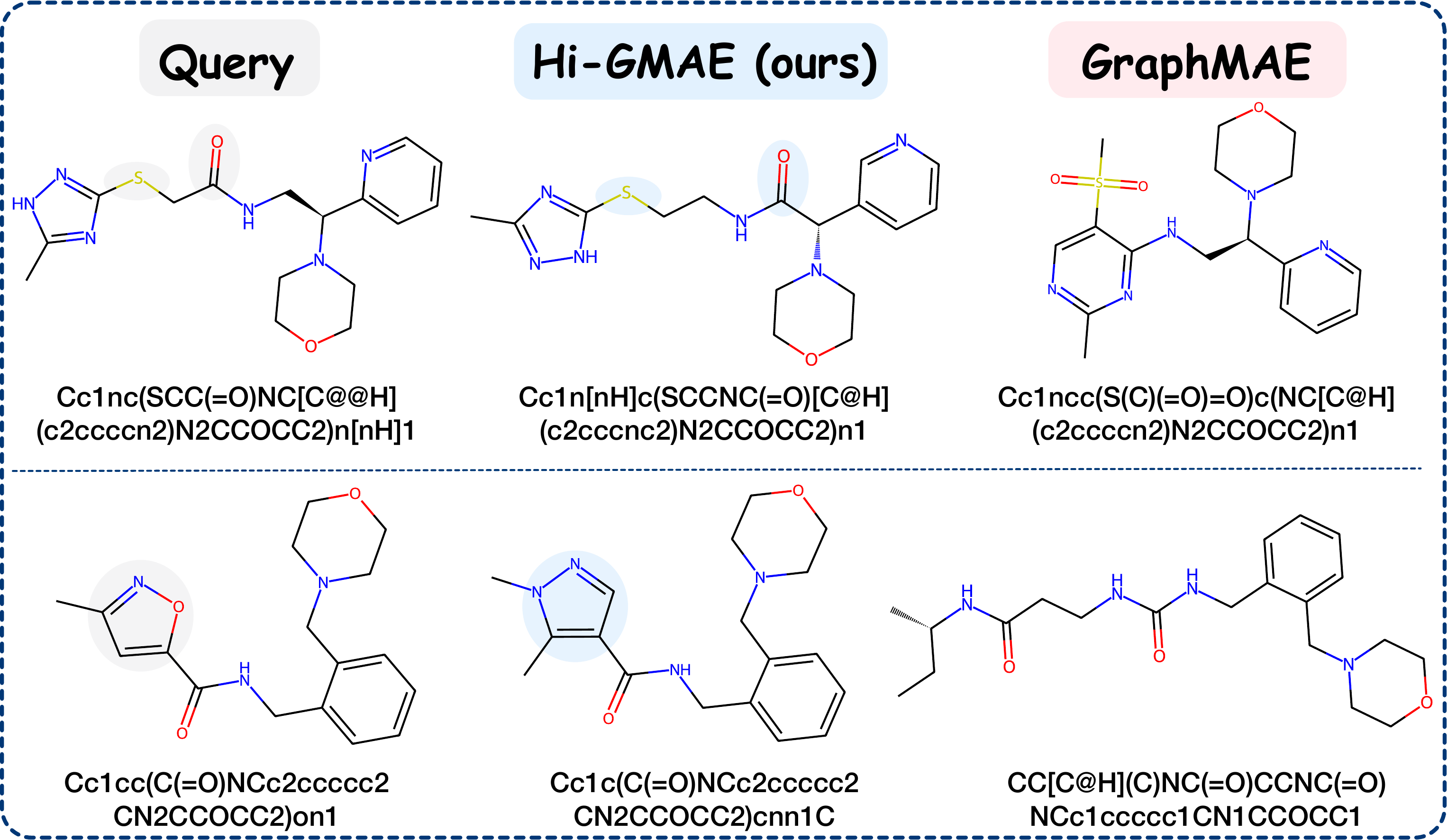}
\end{center}
\caption{Visualization of the top-ranked molecule, identified by molecular representation similarity, to the query molecule from ZINC15. The molecule representations obtained from the pre-trained model in the transfer learning task. }
\label{fig:molecular}
\end{figure}

For a deeper exploration of the semantic information embedded in the pre-trained representations generated by Hi-GMAE (trained on the ZINC15 dataset), we assess the similarity (\textit{i.e.}, cosine similarity) between a given query molecule and other molecules, displaying the top-most similar molecules. Figure~\ref{fig:molecular} illustrates that the molecule ranked highest by Hi-GMAE (Top@1) exhibits remarkable similarity to the query molecule, mirroring both the types of nodes and the overall graph structure. In contrast, the Top@1 molecule identified by GraphMAE~\cite{graphmae} only shares atom types with the query molecule, significantly lacking in replicating essential structural features such as functional groups (\textit{e.g.}, \ce{[=O]} and \ce{SCC}). These findings highlight Hi-GMAE's enhanced capabilities in capturing complex graph structures, indicating its superior efficacy in graph learning tasks.

\end{document}